\documentclass[11pt]{article}
\usepackage[T1]{fontenc}
\usepackage{graphicx, subcaption}
\usepackage{natbib}
\PassOptionsToPackage{numbers, compress}{natbib}
\usepackage[utf8]{inputenc} 
\usepackage[T1]{fontenc}    
\usepackage{hyperref}       
\usepackage{url}            
\usepackage{booktabs}       
\usepackage{amsfonts, bm}       
\usepackage{nicefrac}       
\usepackage{microtype}      
\usepackage{xcolor}         
\usepackage{algorithm}
\usepackage{graphicx}
\usepackage[noend]{algpseudocode}
\usepackage{dsfont}
\usepackage{amsmath}
\usepackage[english]{babel}
\usepackage{amsthm}
\usepackage{xcolor}

\usepackage{wrapfig}

\usepackage{verbatim}
\usepackage{amsmath}
\usepackage{fullpage}

\newcommand{\red}[1]{{\leavevmode\color{red}{#1}}}

\newcommand{\green}[1]{{\leavevmode\color[RGB]{0,128,0}{#1}}}
\newcommand{\purple}[1]{{\leavevmode\color[RGB]{128,0,255}{#1}}}

\newcommand{\lblue}[1]{{\leavevmode\color[RGB]{0,128,255}{#1}}}

\newcommand\todo[1]{{\red{TODO: {#1}}}}

\newcommand\toref{\red{[REF]}}

\newcommand{\gau}[1]{\green{[Gaurav: {#1}]}}
\newcommand{\vin}[1]{\purple{[Vineet: {#1}]}}
\newcommand{\aur}[1]{\lblue{[Aurghya: {#1}]}}

\newcommand{\SRM}{\texttt{SRM-ALG}}
\newcommand{\CB}{\texttt{CB-ALG}}
\newcommand{\CRM}{\texttt{CRM-ALG}}
\newcommand{\UCB}{\texttt{UCB}}
\newcommand{\MAB}{\texttt{MAB}}
\newcommand{\SR}{\texttt{SR}}
\newcommand{\UE}{\texttt{UE}}

\newcommand{\z}{\mathbf{z}}

	\ifdefined\DEBUG
	    
	    \newcommand{\VIN}[1]{\purple{#1}}
	    \newcommand{\AUR}[1]{\lblue{{#1}}}
	\else
	    
	    \newcommand{\VIN}[1]{#1}
	    \newcommand{\AUR}[1]{#1}
	\fi

\newtheoremstyle{break}
  {\topsep}{\topsep}%
  {\itshape}{}%
  {\bfseries}{}%
  {\newline}{}%
\theoremstyle{break}

\newtheorem{theorem}{Theorem}[section]
\newtheorem{assumption}[theorem]{Assumption}

\newtheorem{lemma}[theorem]{Lemma}
\newtheorem{observation}[theorem]{Observation}

\begin{document}
\title{A Causal Bandit Approach to Learning Good Atomic Interventions in Presence of Unobserved Confounders}
\author{
Aurghya Maiti\\
\footnotesize{Adobe Research}\\
\normalsize{\tt aurgmait@adobe.com}
\and Vineet Nair\footnote{Equal contribution and alphabetical within.}\\
\footnotesize{Technion Israel Institute of Technology}\\
\normalsize{\tt vineet@cs.technion.ac.il}
\and {Gaurav Sinha\footnotemark[\value{footnote}]}\\
\footnotesize{Adobe Research}\\
\normalsize{\tt gasinha@adobe.com}}
\maketitle
\date{}

\begin{abstract}
We study the problem of determining the best intervention in a Causal Bayesian Network (CBN) specified only by its causal graph. We model this as a stochastic multi-armed bandit (MAB) problem with side-information, where the interventions correspond to the arms of the bandit instance. First, we propose a simple regret minimization algorithm that takes as input a semi-Markovian causal graph with atomic interventions and possibly unobservable variables, and achieves $\tilde{O}(\sqrt{M/T})$ expected simple regret, where $M$ is dependent on the input CBN and could be very small compared to the number of arms. We also show that this is almost optimal for CBNs described by causal graphs having an $n$-ary tree structure. Our simple regret minimization results, both upper and lower bound, subsume previous results in the literature, which assumed additional structural restrictions on the input causal graph. In particular, our results indicate that the simple regret guarantee of our proposed algorithm can only be improved by considering more nuanced structural restrictions on the causal graph. Next, we propose a cumulative regret minimization algorithm that takes as input a general causal graph with all observable nodes and atomic interventions and performs better than the optimal MAB algorithm that does not take causal side-information into account. We also experimentally compare both our algorithms with the best known algorithms in the literature. To the best of our knowledge, this work gives the first simple and cumulative regret minimization algorithms for CBNs with general causal graphs under atomic interventions and having unobserved confounders.  
\end{abstract}

\section{Introduction}


Causal Bayesian Networks or CBNs \cite{Pearl00} have become the natural choice for modelling causal relationships in many real-world situations such as Price-Discovery \cite{Haigh2004}, Computational-Advertising \cite{Bottou2013}, Healthcare \cite{Velikova2014}, etc. 
\VIN{A CBN has two components: a directed acyclic graph (DAG) called the causal graph, and a joint probability distribution over the random variables labelling the nodes of the graph. The edges in the DAG of a CBN represent direct causal relationships and therefore it captures the data generation process.} 
In its most general setup, only a subset of the variables appearing in the CBN are observable and the rest are considered hidden \VIN{(see Definition $1.3.1$ in \cite{Pearl00})}. 
CBNs enable modelers to simulate the effect of external manipulations called \emph{interventions}, wherein, observable variables are forcibly fixed to certain desired values. Such an intervention on a CBN is performed via the $do()$ operator, which breaks incoming edges into variables being intervened on and fixes their values as desired. Data generated from the resulting model is the simulated outcome of the intervention. In the presence of an outcome variable of interest $Y$ (assumed to be observable), a natural question is to find the intervention which maximizes the expected value of this outcome. A simple albeit \VIN{practically interesting and also technically challenging} version of this problem is when interventions manipulate only a single variable (also known as \emph{atomic interventions}), as this amounts to determining the variable that has the highest causal impact on $Y$.
The problem of learning the best atomic intervention mentioned above was formulated as a sequential decision making problem called \emph{Causal Bandits} (CB) in \citet{LattimoreLR16}. In CB,  access to the underlying DAG of the CBN is assumed but the associated conditional probability distributions are unknown. The outcome variable $Y$ is considered as a reward variable and the set of allowed atomic interventions are modelled as arms of a bandit instance. In addition, there is an \emph{observational arm} corresponding to the empty intervention, and pulling the observational arm generates a sample from the joint distribution of all observable variables. Here, identifying the best atomic intervention is equivalent to the well-studied \emph{best-arm identification} problem in a multi-armed bandit (\MAB) instance. However, while pulling an arm in CB an algorithm has access to \emph{causal} side information derived from the causal graph associated with the input CBN. See \cite{LattimoreLR16} and the references therein for a comparison of CB and \MAB\ problems with other types of side-information. \\

\begin{wrapfigure}{r}{0.4\textwidth}
     \centering
     \includegraphics[scale=0.6]{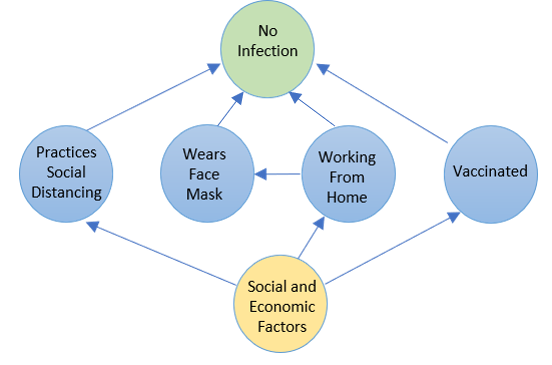}
     \caption{Causal Graph: Infection Prevention}
     \label{fig:exampleGraph}
\end{wrapfigure}
We now provide a motivating example where determining the best atomic intervention is important. Suppose a policy-maker is required to identify the best precautionary measure that should be enforced to reduce spread of a disease. The available measures are mandating social distancing, wearing of face mask, making people work from home and preventive vaccinations. Since the effect of each measure needs to be isolated while disrupting public life minimally, the policy-maker can enforce at most one of these measures at a given time.
The policy-maker can conduct surveys to collect data from public about which measures were taken by them (other than the one enforced) and whether they got infected or not. The goal would then be to design a mechanism of implementing such enforcement one by one, during a time period and collecting the respective survey data to identify the best measure to enforce. Note that, using domain knowledge of health experts, policy makers can have access to an underlying causal graph such as the one in Fig. \ref{fig:exampleGraph}. They would want to use this graph to make better decisions of if and when to enforce a particular measure during the course of their investigation.\\

In this work, we study CB for general causal graphs with atomic interventions and identifiable unobserved confounders (UCs). These are unobserved variables that directly affect more than one node in the causal graph. For eg. in Fig. \ref{fig:exampleGraph}, social and economic background of a person affects their chances of following different precautionary measures, but these factors are extremely difficult to observe in survey data. To the best of our knowledge, this is the first work that analyses the regret of causal bandit algorithms for general causal graphs in the presence of UCs. \citet{YabeHSIKFK18} proposed a causal bandit algorithm for general graphs with interventions that could simultaneously manipulate any number of variables. But the input causal graph is assumed to have no UCs, and the performance of the proposed algorithm is equivalent to that of the best \MAB\ algorithm for atomic interventions.
\subsection{Our Contributions}
We study CB with respect to two standard objectives in \texttt{MAB}: simple and cumulative regret. The simple regret captures the best arm identification setting described above, whereas the cumulative regret is more natural when the the agent is required to maximize the cumulative reward at the end of $T$ rounds instead of determining the best arm. We state our contributions below; the meanings of the relevant terminologies are defined in Sec. \ref{sec: model and prelim}. \\

\textbf{Simple Regret Minimization}: We propose a simple regret minimization algorithm called \SRM. Given a semi-Markovian causal graph with atomic interventions, \SRM\ attains $\Tilde{O}(\sqrt{M/T})$ expected simple regret (see Theorem \ref{theorem: UB-SR}). Here, $M$ depends on the input CBN and $M \leq N$, where $N$ is the number of intervenable nodes.\footnote{We note that the algorithm only receives a causal graph as input and the associated probability distribution is unknown  to the algorithm.} In Sec. \ref{sec: simple regret for general graphs} we give examples of graphs, where $M \ll N$, and hence \SRM\ performs better than a standard bandit algorithm which achieves $\Omega(\sqrt{N/T})$ expected simple regret (Thm. 4 in \cite{AudibertBM10}). \\

We note that \citet{LattimoreLR16} and \citet{NairPS21} propose algorithms where the causal graph for the input CBN is a parallel graph and a no-backdoor graph, respectively.\footnote{These graphs have no-backdoor paths from an intervenable node to $Y$ and hence the conditional distributions are same as interventional distributions.} 
For these special classes of graphs, \SRM\ recovers the regret guarantees given in \cite{LattimoreLR16,NairPS21}. Hence, \SRM\ can be viewed as a \emph{significant and non-trivial} generalization of these algorithms to general graphs with atomic interventions and identifiable unobserved confounders.
As noted earlier, the regret guarantee of the causal bandit algorithm in \cite{YabeHSIKFK18} for general causal graphs with atomic interventions is $\tilde{O}(\sqrt{N/T})$,
and in particular its performance is not better than the optimal \MAB\ algorithm that does not take into account the causal side-information. In Sec. \ref{sec: experiments}, we experimentally compare the regret guarantee of \SRM\ with the algorithm in \cite{YabeHSIKFK18}, as well as \MAB\ algorithms that does not take into account the causal side-information.\\ 

\textbf{Lower Bound on Simple Regret}: Next, we show that \SRM\ is almost optimal for CBNs associated with a large and important class of causal graphs. Specifically, in Theorem \ref{theorem: LB-Tree}, we show that for any causal graph which is an $n$-ary tree with $N$ nodes (all intervenable), and an $M\in [1,N]$ there is probability distribution compatible with the the causal graph such that the expected simple regret of any algorithm at the end of $T$ rounds is $\Omega(\sqrt{M/T})$. We remark that these graphs naturally capture important CBNs like causal trees \cite{GreenewaldKSMKA19}. Also, the class of graphs considered in Theorem \ref{theorem: LB-Tree} subsumes the parallel graph model, and the lower bound for parallel graphs in Theorem \ref{theorem: LB-Tree} matches the lower bound for parallel graphs given in \cite{LattimoreLR16}. Importantly, Theorem \ref{theorem: LB-Tree} implies that the regret guarantee of \SRM\ can be only improved by considering more nuanced structural restrictions on the causal graph describing the input CBN, which could enable more causal information sharing between the interventions. \\

\textbf{Cumulative Regret Minimization}: For general causal graphs with all observable nodes and atomic interventions, we propose \CRM\ that achieves constant expected cumulative regret if the observational arm is optimal, and otherwise achieves better regret than the optimal \texttt{MAB} algorithm that does not take into account the causal side-information (see Theorem \ref{theorem: UB-CRM}). Cumulative regret minimization in general graphs were also studied by \citet{LU2020} and \citet{NairPS21}. However, they crucially assume that the distribution of the parents of the reward node is known for every intervention, which is limiting in practice. Our algorithm (\CRM) and its analysis do not make this assumption.\\
\subsection{Related Work} 
As noted before, causal bandits was introduced in \cite{LattimoreLR16}, where  an almost optimal algorithm was proposed for CBNs associated with a parallel causal graph. Recently, a similar algorithm for simple regret minimization along with an algorithm for cumulative regret minimization was proposed for no-backdoor graphs in \cite{NairPS21}, and the observation-intervention trade-off was studied when interventions are costlier than observations. An importance sampling based algorithm was proposed by \citet{SenSDS17} to minimize simple regret but only soft-interventions at a single node were considered. The cumulative regret minimization problem for general causal graphs was studied in \cite{LU2020,NairPS21}, but they assume the knowledge of the distributions of the the parents of the reward variable for every intervention. 
\citet{SenSKDS17} studied the contextual bandit problem where the observed context influences the reward via a latent confounder variable, and proposed an algorithm with better guarantee compared to standard contextual bandit. \cite{LeeB18,LeeB19} gave a procedure to compute the minimum possible intervention set by removing sub-optimal interventions identifiable from the input causal graph, and they empirically demonstrated that ignoring such information leads to huge regret. Recently, \cite{lu2021causal} introduced the causal Markov decision processes, where at each state a causal graph determines the action set, and gave algorithms that achieve better policy regret when the causal side-information is taken into account. Finally, in a different line of work \citet{BareinboimFP15} established that in the presence of UCs determining the intervention that has maximum reward may not be always desirable. 


\section{Model and Preliminaries}\label{sec: model and prelim}
\textbf{Causal Bayesian Network}: A CBN is a tuple $\mathcal{C} = (\mathcal{G},\mathbb{P})$, where $\mathcal{G} = (\mathbf{V},\mathbf{E})$ is a directed acyclic graph called the causal graph, and $\mathbf{V}$ and $\mathbf{E}$ are the set of nodes and edges in $\mathcal{G}$ respectively. The nodes in $\mathbf{V}$ are labelled by random variables, and $\mathbb{P}$ is the joint distribution over $\mathbf{V}$ that factorizes over $\mathcal{G}$. In a CBN, certain nodes are not observable and are termed hidden/unobservable variables and denoted $\mathbf{U} \subseteq \mathbf{V}$. A node $V_i$ is called the parent of $V_j$ and $V_j$ the child of $V_i$, if there is a directed edge from $V_i$ to $V_j$ in $\mathbf{E}$. The set of \emph{observable} parents of a node $V$ is denoted as $\mathbf{Pa}(V)$. The in-degree of a node is the maximum number of directed edges entering the node, and the in-degree of a causal graph is the maximum of the in-degrees of its nodes.  
An intervention on node $X \in \mathbf{V}$ is denoted as $do(X=x)$, where $X$ is set as $x \in \{0,1\}$ and all the edges from the parents of $X$ to $X$ are removed, and the resulting graph defines a probability distribution $\mathbb{P}(\mathbf{V}\setminus \{X\} \mid do(X=x))$ over $\mathbf{V}\setminus \{X\}$. The intervenable nodes are denoted $\mathbf{X} \subseteq \mathbf{V}$. \\

\textbf{Causal Bandits}: A causal bandit algorithm receives as input a CBN $\mathcal{C} = (\mathcal{G}, .)$, the associated set of intervenable nodes $\mathbf{X} \subseteq \mathbf{V}$ and the designated reward node $Y\in \mathbf{V}$; in particular the algorithm only receives as input the causal graph $\mathcal{G}$, but the $\mathbb{P}$ associated with $\mathcal{C}$ is unknown to the algorithm. 
 We assume there are $N$ intervenable nodes $\mathbf{X} = \{X_1, \ldots, X_N\}$, and there are $2N$ interventions denoted $a_{i,x} = do(X_i =x)$ for $i\in [N]$ and $x\in \{0,1\}$. The empty intervention $do()$ is denoted as $a_0$. These $2N+1$ interventions constitute the arms $\mathcal{A}= \{a_{i,x} \mid i \in [N], x\in \{0,1\}\} \cup \{a_0\}$ of the bandit instance. A causal bandit algorithm is a sequential decision making process that at each time $t$, makes an intervention $a_t \in \mathcal{A}$, and observes the sampled values of the nodes in $\mathbf{V}\setminus \mathbf{U}$ including the value of the node $Y$; the values of the nodes in $\mathbf{U}$ are unobservable. The values of nodes $V \in \mathbf{V}$, $X \in \mathbf{X}$ and $Y$ sampled at time $t$ are denoted as $V_t, X_t$, and $Y_t$ respectively. Throughout the paper we use $i, x$, and $a$ to index the sets $[N]$, $\{0,1\}$, and $\mathcal{A}$ respectively. The expected reward corresponding to intervention $a_{i,x}\in \mathcal{A}$ and $a_0 \in \mathcal{A}$ is denoted as $\mu_{i,x} = \mathbb{E}[Y\mid do(X_i =x)]$ and $\mu_0 = E[Y]$. We study the causal bandit problem with respect two standard objectives in bandit literature: simple and cumulative regret. \\

\textbf{Simple Regret}: The expected simple regret of an algorithm $\texttt{ALG}$ that outputs arm $a_T$ at the end of $T$ rounds is defined as $r_{\texttt{ALG}}(T) = \max_{a\in \mathcal{A}} \mu_a - \mu_{a_T}$.\\

\textbf{Cumulative Regret}: Let \texttt{ALG} be an algorithm that plays arm $a_t$ at time $t\in [T]$. Then the expected cumulative regret of $\texttt{ALG}$ at the end of $T$ rounds is defined as $R_{\texttt{ALG}}(T) = \max_{a\in \mathcal{A}} \mu_a\cdot T - \sum_{t\in [T]}\mu_{a_t}$.\\

\textbf{Model Assumptions}: Throughout this paper we make the following assumptions on the input CBN. \VIN{These assumptions are minimal compared to the previous works \cite{LattimoreLR16, LU2020, NairPS21}. Also, Assumptions 2 and 3 is in some sense necessary to prove theoretical guarantees in the presence of UCs.}\\

1. The distribution of any intervenable node $X_i$ conditioned on its parents $\mathbf{Pa}(X_i) =\mathbf{z}$ is Bernoulli with parameter $p_{i,\z}$; that is $\mathbb{P}(X_i = 1\ \mid \mathbf{Pa}(X_i) =\z\ ) = p_{i,\z}$. \\ 

2.  The causal graph $\mathcal{G}$ corresponding to the input CBN is semi-Markovian, that is a hidden variable in $\mathbf{U}$ does not have parents and is a parent of at most two observable variables in $\mathbf{V}\setminus \mathbf{U}$. \\

3. First perform the following transformation in $\mathcal{G}$: if there is a hidden variable $U \in \mathbf{U}$ that is a common parent of $V_i$ and $V_j$ then add a bi-directed edge between $V_i$ and $V_j$ in $\mathcal{G}$. In the transformed graph, we assume there does not exist a path of bi-directed edges from an intervenable node $X$ to a child of $X$. \VIN{This assumption known as \emph{identifiability} is necessary and sufficient for the estimation of the distribution associated with a given causal graph \cite{TianP02}.}\\


\textbf{Preliminaries}: We use $\mathbf{z}$ to denote a realization of $\mathbf{Pa}(X_i)$ for some $i$, in particular $\mathbf{z} \in \text{Domain}(\mathbf{Pa}(X_i))$. As noted previously, the presence of an unobserved common-parent $U$ of $V_i$ and $V_j$ is denoted by a bi-directed edge between $V_i$ and $V_j$, and $U$ is removed from the input causal graph. The resulting graph with both directed and bi-directed edges is referred to as an \emph{acyclic directed mixed graph} (ADMG). Further,  a \emph{c-component} in an ADMG is a subset of $\mathbf{V}$ connected only via bi-directed edges, that is $V_1$ and $V_2$ belong to the same c-component if and only if there is path in the graph comprising only of bi-directed edges. It is easy to see that $\mathbf{V}$ can be partitioned into disjoint c-components.\footnote{If a node is not incident by any bi-directed edge then its c-component is itself.} 
The c-component containing the intervenable node $X_i$ is denoted $S_i$, and let $k_i = |S_i|$, and $d_i$ be the number of observable parents of $X_i$, that is $|\mathbf{Pa}(X_i)| = d_i$. Throughout the paper, $k = \max_{i\in [N]} k_i$, and $d$ is the in-degree of the input causal graph. Finally for an intervenable node $X_i$, let $\mathbf{Pa}^{+}(S_i) = S_i \cup \bigcup_{V \in S_i} \mathbf{Pa}(V)$, 
and let $\mathbf{Pa}^c(X_i) = \mathbf{Pa}^{+}(S)\setminus X_i$.

\section{Simple Regret for General Graphs}\label{sec: simple regret for general graphs}
\label{simple-regret}
In this section, we state and analyze our simple regret minimization algorithm called \SRM. 
Our proposed algorithm repeatedly plays the observational arm $a_0$ for the first $T/2$ rounds, and estimates the rewards corresponding to each intervention from the samples corresponding to the observational arm. 
This step is accomplished by adapting a procedure from \cite{BhattacharyyaGKMV20} which efficiently estimates distributions resulting from an atomic intervention using observational samples. We remark that previous works in \cite{LattimoreLR16} and \cite{NairPS21} imposed structural restrictions on the input causal graphs which allowed observational samples to be  directly used for estimating rewards corresponding to interventions\footnote{The restrictions in these work ensured that the conditional distributions are equal to the corresponding do distributions}. \SRM\ even in the case of general causal graphs with hidden variables, is able to efficiently estimate the rewards of all the arms simultaneously using the observational arm pulls.
%
The quality of the reward estimate computed at the end of $T/2$ rounds for an arm depends on the unknown $\mathbb{P}$, and are not equally good.
The main technical challenge in the next part is to identify the optimal number of arms with bad estimates to intervene upon so as to minimize the worst-case expected simple regret. 
The algorithm efficiently identifies such arms, and the remaining $T/2$ rounds are equally partitioned among them.  Next, we give the the main technical ideas used in \SRM.  
\begin{algorithm}
\caption{\SRM: (Best Arm Identification in General CBN)} \label{SR-algorithm}
\begin{algorithmic}
\State INPUT: Causal graph $\mathcal{G}$, and the set of intervenable nodes $\mathbf{X}\subseteq \mathbf{V}$.
\end{algorithmic}
\begin{algorithmic}[1]
\setlength{\lineskip}{5pt}
\State $\mathsf{His} = \{\}$ ~~\textcolor{gray}{/* $\mathsf{His}$ would be used to keep the history of sampled values in the first $T/2$ rounds. */}
\For {$t \in [1, \ldots , T/2]$}
    \State Play arm $a_0$ and let $\mathsf{His} =\mathsf{His} \cup  \{\mathbf{V_t}\setminus \mathbf{U}_t, Y_t\}$.
\EndFor
\State For each $(i,x)$, use Algorithm \ref{alg: estimating rewards from observations} in Appendix \ref{secappendix: estimating rewards from observations} with input $\mathcal{G},\mathcal{H}$ to compute $\hat{\mu}_{i,x}$. 
\State For each $(i,x)$, compute $\hat{q}_{i} = (2/T)\cdot \min_{\mathbf{z},x} \{ \sum_{t=1}^{T/2} \mathds{1}\{X_{i,t} = x, (\mathbf{Pa}^c(X_i))_t = \mathbf{z}\} \}$.
\State Compute $\widehat{m}$ as an estimate of $m$, by using $\widehat{q}_{i}$ in place of $q_{i}$.
\State Let $\mathcal{Q} = \{a_{i,x} \in A: \hat{q}_{i}^{k_i} < 1/\widehat{m}\}$.
\For {$a_{i,x} \in \mathcal{Q}$}
    \State Play arm $a_{i,x}$ and observe $Y_t$ for $\frac{T}{2|\mathcal{Q}|}$ rounds.
    \State Re-estimate $\hat{\mu} = \frac{2|\mathcal{Q}|}{T} \sum_{t=1}^{T/2|\mathcal{Q}|} Y_t$.
\EndFor
\State Return estimated optimal $a_T^* \in \arg\text{-}\max_{a \in \mathcal{A}} \hat{\mu}_a$.
\end{algorithmic}
\end{algorithm}

\emph{Steps 1--4}: 
The observational samples collected in first $T/2$ rounds are used to compute the estimates for each arm $a_{i,x}$ at Step 4. The procedure to compute the estimates $\mu_{i,x}$ is given in Algorithm \ref{alg: estimating rewards from observations}, Appendix \ref{secappendix: estimating rewards from observations}. We note that Assumptions 2 and 3, stated in Sec. \ref{sec: model and prelim} are required to compute the estimates at Step 4. 
We note that the quality of the estimate $\widehat{\mu}_{i,x}$ computed at Step 4 depends on the joint probability of the nodes in the c-component to which $X_i$ belongs and their parents, and is in particular bad if this is low. \\

\emph{Steps 5--11}: At Steps 5 and 6, \SRM\ determines the number of arms with bad estimates that could be intervened upon in the remaining rounds so as to minimize regret. This is done by estimating the quantity $m(\mathcal{C})$ defined next; the meaning of relevant notations can be found in Sec. \ref{sec: model and prelim}. Let $q_{i} = \min_{\mathbf{z},x} \mathbb{P}(X_i = x, \mathbf{Pa}^c(X_i) = \mathbf{z})$, and $\mathbf{q} = \{q_{i} : 1\leq i \leq N\}$.\footnote{In the absence of unobserved variables, we only consider parents on the backdoor paths.} Further, let $I_\tau = \{i : q_{i}^{k_i} < 1/\tau\}$ for $\tau \in [2,2N]$. Then $m(\mathcal{C}) = \min \{\tau : |I_\tau| \leq \tau \}$. When $\mathcal{C}$ is immediate from the context, we omit $\mathcal{C}$ and just use $m$. Note that $m$ is a function that takes as input a CBN $\mathcal{C}$, and returns a value in $[2,N]$. Also, the value of $m$ depends only on $\mathbf{q}$ and $k_i$ for $i \in [N]$. Since $\mathbb{P}$ is not known to \SRM, $\mathbf{q}$ is also not known and is estimated at Step 5. We note that our definition of $m$ coincides with the definitions of $m$ given for parallel graphs and no-backdoor graphs in \cite{LattimoreLR16} and \cite{NairPS21} respectively, and the regret guarantee of \SRM\ for these graphs are the same as those of the respective algorithms in these works. \\

The function $m$ is defined to determine the optimal trade-off between the number of arms to be pulled in the remaining $\frac{T}{2}$ rounds and the quality of $\widehat{\mu}_{i,x}$ computed at Step 4 so as to minimize expected regret. In particular, $m$ determines the optimal $I_\tau$, that trades-off the number of arms with bad estimates with the ones that have bad estimates but are not part of $I_\tau$. We show that the expected simple regret of \SRM\ in Theorem \ref{theorem: UB-SR} stated below is $\tilde{O}(\sqrt{m/T})$, which is an instance-dependent regret guarantee as $m$ depends on the input CBN. If $m \ll N$ then \SRM\ performs better than the optimal \texttt{MAB} algorithm. In particular, \SRM\ needs to explore only \AUR{$2m$} arms after $T/2$ rounds compared to the $2N$ actions that must be explored by a standard best-arm identification \texttt{MAB} algorithm which achieves $\Omega(N/T)$ expected simple regret. It is easy to see that there are CBNs with $m \ll N$; for example consider a CBN $\mathcal{C} = (\mathcal{G},\mathbb{P})$ with $N$ intervenable nodes and in-degree at most $k-1$, and let $k$ be such that $2^k \ll N$. Further, let $\mathbb{P}$ be such that for at most $2^k$ nodes, chosen in the reverse topological order, the conditional probability of a node given its parents is Bernoulli with parameter $1/2^{k+1}$, and for the remaining nodes the conditional probability of a node given its parents is Bernoulli with parameter $1/2$. Now it is easy to see that $m(\mathcal{C}) \leq 2^k \ll N$. The proof of Theorem \ref{theorem: UB-SR} is given in Appendix \ref{secappendix: proof of SRM}.



\begin{theorem} \label{theorem: UB-SR}
The expected simple regret of \SRM\ at the end of $T$ rounds is $r_{\SRM}(T) = O\bigg(\sqrt{\frac{m(\mathcal{C})}{T}\log \frac{NT}{m(\mathcal{C})}}\bigg)$, where $\mathcal{C}$ is the input CBN.
\end{theorem}
\section{Lower Bound for Simple Regret}\label{sec: lower bound for general graphs}
A closer inspection of \SRM\ given in Sec. \ref{sec: simple regret for general graphs} reveals that the algorithm only leverages the causal side-information available while pulling the observational arm. Hence, there remains a possibility that a better algorithm than \SRM\ could be designed which uses the information shared between any two interventions to achieve a better regret guarantee. In this section, we show that this is not possible for a large and important class of causal graphs that we call tree-graphs and denote it as $\mathsf{T}$. 
Each graph in $\mathsf{T}$ is an $n$-ary tree, where each node can have $2$ to $n$ children. Additionally, all the leaves in the graph are connected to the outcome node $Y$, and every node except the outcome node. We also assume that all the nodes of a graph in $\mathsf{T}$ is observable, which from the perspective of lower bound is desirable. 
Note that although a causal bandit algorithm receives as input a CBN $\mathcal{C}$, it only sees the corresponding causal graph $\mathcal{G}$ of the CBN and the associated distribution $\mathbb{P}$ is unknown to the algorithm. Since there are multiple probability distributions that are compatible with a given $\mathcal{G}$ the algorithm is required to learn the unknown $\mathbb{P}$ through the arm pulls. We show in Theorem \ref{theorem: LB-Tree} that for a given causal graph $\mathcal{G}$ in $\mathsf{T}$ and a positive integer $M$, any algorithm must explore at least $O(M)$ arms to learn $\mathbb{P}$ with sufficient confidence to minimize the worst-case expected simple regret.


\begin{theorem} \label{theorem: LB-Tree}
Corresponding to every causal graph $\mathcal{G} \in \mathsf{T}$, with $N$ intervenable nodes and a positive integer $M \leq N$, there exists a probability measure  $\mathbb{P}$ and CBN $\mathcal{C} = (\mathcal{G}, \mathbb{P})$ such that $m(\mathcal{C}) = M$ and the expected simple regret of any causal bandit algorithm \texttt{ALG} is $r_{\texttt{ALG}}(T) = \Omega\big(\sqrt{m(\mathcal{C})/T} \big)$.
\end{theorem}
The proof of Theorem \ref{theorem: LB-Tree} is in Appendix \ref{secappendix: proof of lower bound for tree}. 
Recall, we had noted in Sec. \ref{sec: simple regret for general graphs} that $m$ for a $\mathcal{C}$ was completely defined by $\mathbf{q}$ and $\mathcal{G}$; in particular the definition of $m$ does not depend on the entire probability distribution corresponding to $\mathcal{C}$. We conclude this section by showing in Theorem \ref{theorem: LB-given-q} that the dependence of the regret on the quantity $\mathbf{q}$ in the definition of $m$ is optimal for certain graphs. In Theorem \ref{theorem: LB-given-q}, a $\mathbf{q}$ is valid if there exists a probability measure $\mathbb{P}$ for the given the graph $\mathcal{G}$, which results in the given $\mathbf{q}$. The proof of Theorem \ref{theorem: LB-given-q} is in Appendix \ref{secappendix: lower bound for tree given q}.
\begin{theorem} \label{theorem: LB-given-q}
There exists a fully observable causal graph $\mathcal{G}$ with \AUR{$N \geq 3$ nodes} such that given a valid $\mathbf{q}$ corresponding to 
$\mathcal{G}$ there is a probability measure $\mathbb{P}$ conforming with $\mathbf{q}$ and CBN $\mathcal{C} = (\mathcal{G},\mathbb{P})$ for which  expected simple regret of any causal bandit algorithm $\Omega\big(\sqrt{m(\mathcal{C})/T} \big)$.
\end{theorem}

\section{Cumulative Regret in General Graph}\label{sec: cumulative regret}
\label{cumulative-regret}
In this section, we propose \CRM, an algorithm to minimize the cumulative regret for a CBN $\mathcal{C} = (\mathcal{G},.)$, where $\mathcal{G}$ is a general causal graph with atomic interventions. Though unlike in \SRM, here we assume all nodes in $\mathcal{G}$ are observable. We also assume $\mathbb{P}(X_i = x, \mathbf{Pa}(X_i) = \mathbf{z}) > 0$ for all $i,x$ and $\mathbf{z}$. \CRM\ is based on the well-known upper confidence bound (\texttt{UCB}) algorithm \cite{AuerCF02}. Similar to the \texttt{UCB} family of algorithms \CRM\ maintains UCB estimates at each round and pulls the arm with the highest \texttt{UCB} estimate. \CRM\ performs better than a standard MAB algorithm by leveraging via \emph{backdoor criteria} the available causal side-information while pulling an observational arm; to compute the UCB estimate of an arm $a_{i,x}$ \CRM\ uses the samples from the observational arm pulls in addition to the samples from the arm pulls of $a_{i,x}$. \CRM\ pulls the observational arm a pre-specified number of times, which ensures a good trade-off between the simultaneous exploration of all the arms obtained while pulling the observational arm and the possible loss in reward. We note that \texttt{CRM-NB-ALG} proposed for no-backdoor graphs in \cite{NairPS21}, also ensures that the observational arm $a_0$ is pulled a pre-specified number of times, but \CRM\ differs from \texttt{CRM-NB-ALG} on how the UCB estimates for the arms are computed at the end of each round. 
Next, we present the details of \CRM.  

\begin{algorithm}
\caption{\CRM\ (Minimizing cumulative regret in general causal graph)} \label{CR-algorithm}
\begin{algorithmic}
\State INPUT: Causal graph $\mathcal{G}$ and the set of intervenable nodes $\mathbf{X}\setminus \mathbf{V}$
\end{algorithmic}
\begin{algorithmic}[1]
\State Pull each arm once and set $t = 2N+2$
\State Let $\beta = 1$
\For {$t = 2N+2, 2N+3, \ldots$}
    \If {$N_{t-1}^0 < \beta^2 \log t$}
        \State Pull $a_t = a_0$
    \Else 
        \State Pull $a_t = \arg\text{-}\max_{a \in A} \bar{\mu}_a(t-1)$
    \EndIf
    \State $N_t^a = N_{t-1}^a + \mathds{1}\{a_t = a\}$
    \State Update $\widehat{\mu}_a(t)$ and $\bar{\mu}_a(t)$ for all $a \in A$ according to Equations \ref{equation: emprical estimate for arm 0 modified}, \ref{equation: emprical estimate for arm i,x modified} and \ref{equation: UCB estimate for arms}.
    \State Let $\widehat{\mu}^* = \max_a \widehat{\mu}_a(t)$
    \If {$\widehat{\mu}_0(t) < \widehat{\mu}^*$}
        \State Set $\beta = \min \{\frac{2\sqrt{2}}{\widehat{\mu}* - \widehat{\mu}_0(t)}, \sqrt{\log t}\}$
    \EndIf    
    \State $t = t+1$
\EndFor
\end{algorithmic}
\end{algorithm}

We use $N_t^{i,x}$ and $N_t^0$ to denote the number of times arms $a_{i,x}$ and $a_0$ have been played at the end of $t$ rounds respectively, and further let $a_t$ denote the arm pulled at round $t$. Also, $\widehat{\mu}_{i,x}(t)$ and $\bar{\mu}_{i,x}(t)$ (respectively $\widehat{\mu}_0(t)$ and $\bar{\mu}_0(t)$) denotes the empirical and \UCB\ estimates of the arm $a_{i,x}$ (respectively arm $a_0$) computed at the end of round $T$  respectively. 
At Step 4 \CRM\ checks if the observational arm is pulled at least $\beta^2\log t$ times, and accordingly either plays the observational arm or the arm with the highest \UCB\ estimate. Here the value of $\beta$ is updated as in Steps 11-12 . As noted before, the chosen update for $\beta$ and the corresponding pre-specified number of pulls for arm $a_0$ delicately balances the exploration-exploitation trade-off in expectation. 
The empirical estimate for arm $a_0$ at Step 9 is computed as follows
\begin{equation}\label{equation: emprical estimate for arm 0 modified}
    \widehat{\mu}_0(t) = \frac{1}{N_t^0}\sum_{s=1}^t \mathds{1}\{Y=1, a_s=a_0\}~.
\end{equation}
The empirical estimate for arm $a_{i,x}$ is involved, and as mentioned before is done by leveraging the following backdoor criteria (see Theorem 3.3.2 in \cite{PEARL2009})
$$\mathbb{P}\{Y=1\mid do(X_i =x)\} = \sum_{\mathbf{z}} \mathbb{P}\{Y=1 \mid X_i=x, \ \mathbf{Pa}(X_i)=z\}\mathbb{P}\{\mathbf{Pa}(X_i)=\mathbf{z}\}\ .$$
Let the set of time steps $s \leq t$ at which arm $a_0$ is pulled be denoted by $S_t^0$ and further denote the $j$-th element in $S_t^0$ (ordered according to time) by $t_j^0$. Construct two partitions of this set, one consisting of the odd-numbered elements, $S_{o,t}^0 = \{t_1^0, t_3^0, \dots \}$ and another consisting of even-numbered elements, $S_{e,t}^0 = \{t_2^0, t_4^0, \dots \}$.  We use $S_{o,t}^0$ to estimate $\mathbb{P}\{ Y=1 \mid X_i=x,\ \mathbb{P}\{\mathbf{Pa}(X_i)=\mathbf{z}\}$, and use $S_{e,t}^0$ to estimate $\mathbb{P}\{\mathbf{Pa}(X_i)=\mathbf{z}\}$, for each $\mathbf{z}$. For every possible realization $\mathbf{z}$ of $\mathbf{Pa}(X_i)$, let $S^{i,x}_{\mathbf{z}, t} = \{s\in S_{o,t}^0 \mid X_i = x, \mathbf{Pa}(X_i) = \mathbf{z}\}$, that is $S^{i,x}_{\mathbf{z}, t}$ is the set of time-steps in $S_{o,t}^0$ for which $\{X_i = x, \mathbf{Pa}(X_i) = \mathbf{z}\}$. Further, let $C^{i,x}_t = \min_{\mathbf{z}} |S^{i,x}_{\mathbf{z}, t}|$, and for each $\mathbf{z}$ truncate $S^{i,x}_{\mathbf{z}, t}$ by arbitrarily choosing $C^{i,x}_t$ elements and removing the remaining elements; hence we assume $|S^{i,x}_{\mathbf{z},t}| = C^{i,x}_t$ for all $\mathbf{z}$. Also, denote the $c$-th element of $S^{i,x}_{\mathbf{z},t}$ as $s_{\mathbf{z}, c}^{i,x}$. This truncation ensures that the expectation of the empirical estimate $\widehat{\mu}_{i,x}(t)$, computed as in Equation \ref{equation: emprical estimate for arm i,x modified}, is equal to $\mu_{i,x}$. Partition the set $S_{e,t}^0$ into $C_t^{i,x}$ components, where the $c$-th partition is denoted by $S_{t,c}^{0,i}$ and contains at least $\lfloor|S_{e,t}^0|/C_t^{i,x}\rfloor$ elements. Such a partition can be easily constructed by putting the first $\lfloor|S_{e,t}^0|/C_t^{i,x}\rfloor$ elements into $S_{t,1}^{0,i}$, next $\lfloor|S_{e,t}^0|/C_t^{i,x}\rfloor$ elements into $S_{t,2}^{0,i}$ and so on, with any remaining element being put into the last partition. Compute $\widehat{p}_{t, c}^{\ i, \mathbf{z}} = \sum_{s \in S_{t,c}^{0,i}} \mathds{1}_s\{\mathbf{Pa}(X_i) = \mathbf{z}\}/|S_{t, c}^{0, i}|$, and define the random variable $Y_c^{i,x}$ as follows: Take the $c$-th element of $S_{\mathbf{z},t}^{i,x}$ and multiply by $\widehat{p}_{t,c}^{\ i,\mathbf{z}}$ for all $\mathbf{z}$ and sum them, that is $Y_c^{i,x} = \sum_{\mathbf{z}} \mathds{1}\{Y_{s_{\mathbf{z},c}^{i,x}}=1\}\widehat{p}_{\ t,c}^{i,\mathbf{z}}$. Let $S_t^{i,x}$ be the set of timestamps $s \leq t$, when action taken is $a_{i,x}$, that is $S_t^{i,x} = \{s \in [t] \mid a_s = a_{i,x}\}$. Finally, the empirical estimate $\widehat{\mu}_{i,x}(t)$ of arm $a_{i,x}$ is computed as follows:
\begin{equation}\label{equation: emprical estimate for arm i,x modified}
    \widehat{\mu}_{i,x}(t) = \frac{\sum_{j \in S_t^{i,x}}\mathds{1}\{Y_j=1\} + \sum_{c \in [C_t^{i,x}]} Y_c^{i,x}}{N^{i,x}_t + C^{i,x}_t}
\end{equation}

It is easy to see that the expectation of $\widehat{\mu}_0(t)$ is equal to $\mu_0$, and in Lemma \ref{lemma: unbiased muix} we use the Backdoor Criterion (Section 3.3.1 in \cite{PEARL2009}) to show that $\widehat{\mu}_{i,x}(t)$ is $E[\widehat{\mu}_{i,x}(t)] = \mu_{i,x}$ for every $i,x$. Finally, \CRM\ uses Equations \ref{equation: emprical estimate for arm 0 modified} and \ref{equation: emprical estimate for arm i,x modified} to compute the \UCB\ estimates $\bar{\mu}_{i,x}(t)$ and $\bar{\mu}_0(t)$ of arms $a_{i,x}$ and arm $a_0$ respectively
\begin{equation}\label{equation: UCB estimate for arms}
    \bar{\mu}_{i,x}(t) = \widehat{\mu}_{i,x}(t) + \sqrt{\frac{2 \ln t}{N_t^{i,x} + C^{i,x}_t}}~~~~~\bar{\mu}_0(t) = \widehat{\mu}_0(t) + \sqrt{\frac{2 \ln t}{N_t^0}}
\end{equation}
We bound the expected cumulative regret of \CRM\ in Theorem \ref{theorem: UB-CRM}, where $a* = \arg\text{-}\max_{a \in A} \mu_a$ and for $a\in \mathcal{A}$, $\Delta_a = \mu_{a*} - \mu_a$, $p^{i,x}_{\mathbf{z}} = \mathbb{P}(X_i = x, \mathbf{Pa}(X_i) = \mathbf{z})$, $p_{i,x} = \min_{\mathbf{z}} p^{i,x}_{\mathbf{z}}$. \AUR{Let $Z_i$ be the size of the domain from which $\mathbf{Pa}(X_i)$ takes values and $\eta^{i,x}_T = \max \{0, (1 - Z_i T^{-\frac{p_{i,x}^2}{4}})\}$.}
\begin{theorem}\label{theorem: UB-CRM}
If $a* = a_0$ then at the end of $T$ rounds the expected cumulative regret is \AUR{$O(1)$} 
and otherwise the expected cumulative regret is \AUR{of the order} $\frac{58\ln{T}}{\Delta_0} + \Delta_0 + \sum_{\Delta_{i,x}>0} \Delta_{i,x} \max \bigg(0, 1 + 8\ln{T}\bigg(\frac{1}{\Delta_{i,x}^2} - \frac{p_{i,x} \cdot \eta_T^{i,x}}{36 \Delta_0^2} \bigg) \bigg) + \sum_{\Delta_a > 0} \Delta_a \frac{\pi^2}{3}$.
\end{theorem}
The proof of Theorem \ref{theorem: UB-CRM} is given in Appendix \ref{secappendix: proof of CRM}. Notice that the regret guarantee in Theorem \ref{theorem: UB-CRM} is an instance dependent  constant if $a_0$ is optimal and otherwise slightly better than the \UCB\ family of algorithms. 
Also, it is easy to construct examples of CBNs where the observational arm is optimal, for example see Experiment 3 in Sec. \ref{sec: experiments}.

\section{Experiments}\label{sec: experiments}
In this section, we validate our results empirically. In Experiments 1 and 2 
We compare our proposed algorithm \SRM\ to \VIN{two baseline \MAB\ algorithms: uniform exploration (\UE) and successive rejects (\SR)} \cite{AudibertBM10}. 
\VIN{Although, the regret of \SR\ is $\Omega(\sqrt{N/T})$,} \VIN{in practice \SR\ performs better than other similar \MAB\ algorithms.  We also compare \SRM\ with the simple regret minimization algorithm (Algorithm 3) in \cite{YabeHSIKFK18}. We note that their algorithm runs in time exponential in the in-degree of $Y$, whereas the running time of our algorithm, \SRM, is polynomial in the in-degree of $Y$, and hence a comparison with the algorithm in \cite{YabeHSIKFK18} is only feasible on smaller simpler instance. We provide such a comparison in Appendix \ref{secappendix:additional-experiments}. In Experiment 3, we compare the expected cumulative regret of \CRM\ and \UCB.} \\

\begin{figure}[ht!]
    \centering
    \minipage{0.5\textwidth}
        \includegraphics[width=\columnwidth]{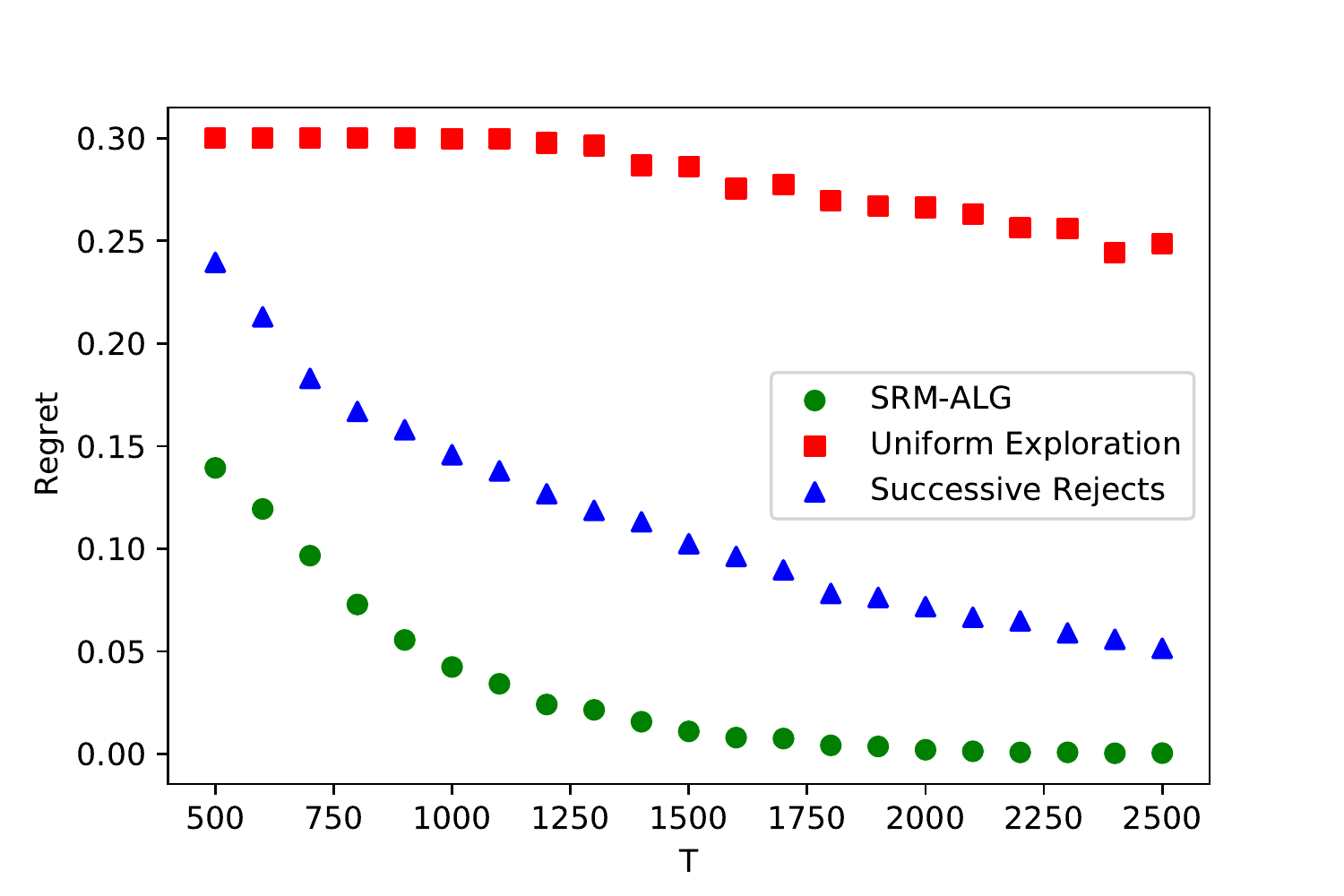}
        \caption{{\small Regret vs Horizon}}
        \label{fig:regret-vs-T}
    \endminipage\hfill
    \minipage{0.5\textwidth}
      \includegraphics[width=\columnwidth]{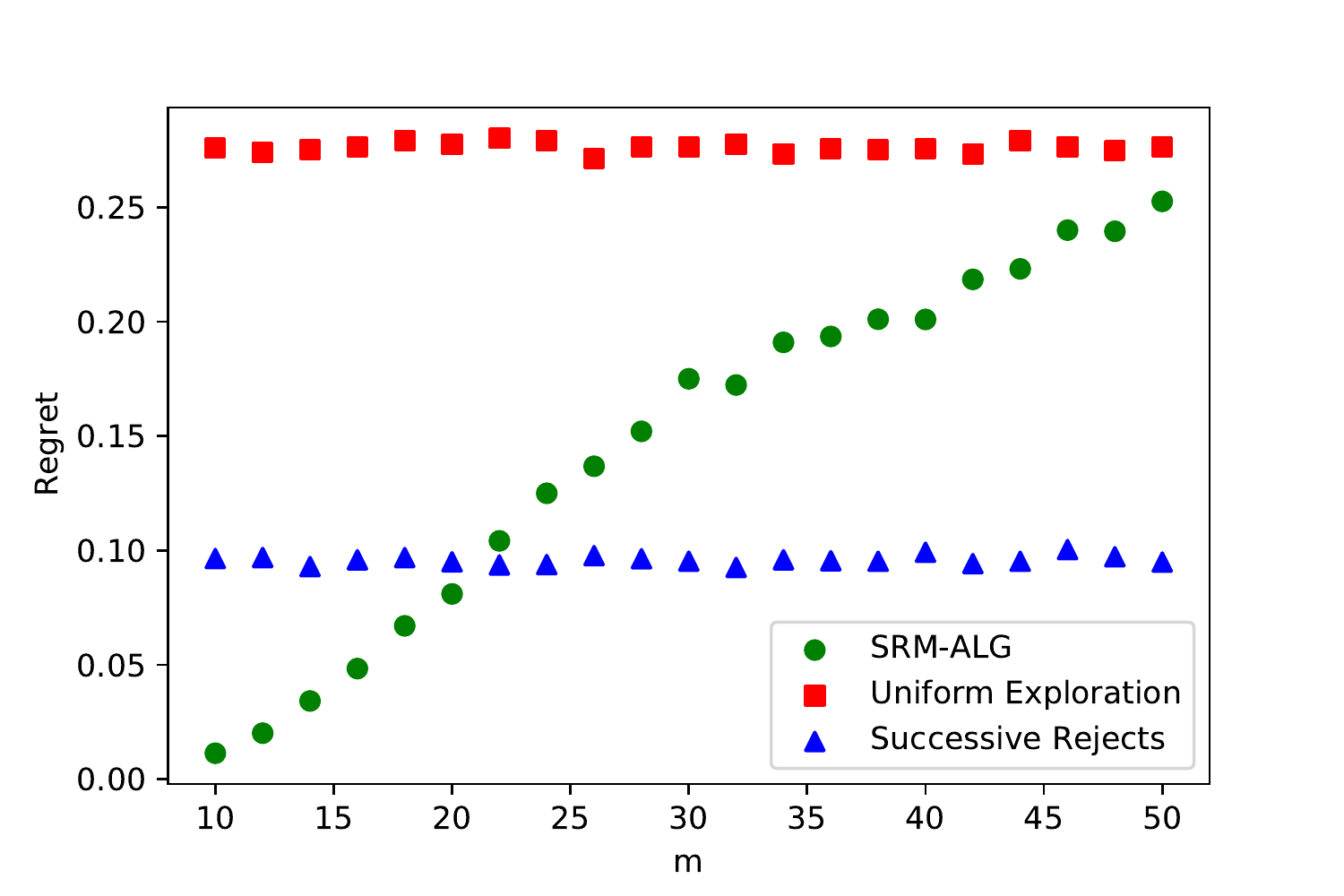}
        \caption[]{{\small Regret vs $m$, $N = 100$}} 	\label{fig:regret-vs-m-100}   
    \endminipage\hfill
\end{figure}
\begin{figure}
\centering
    \minipage{0.5\textwidth}
      \includegraphics[width=\linewidth]{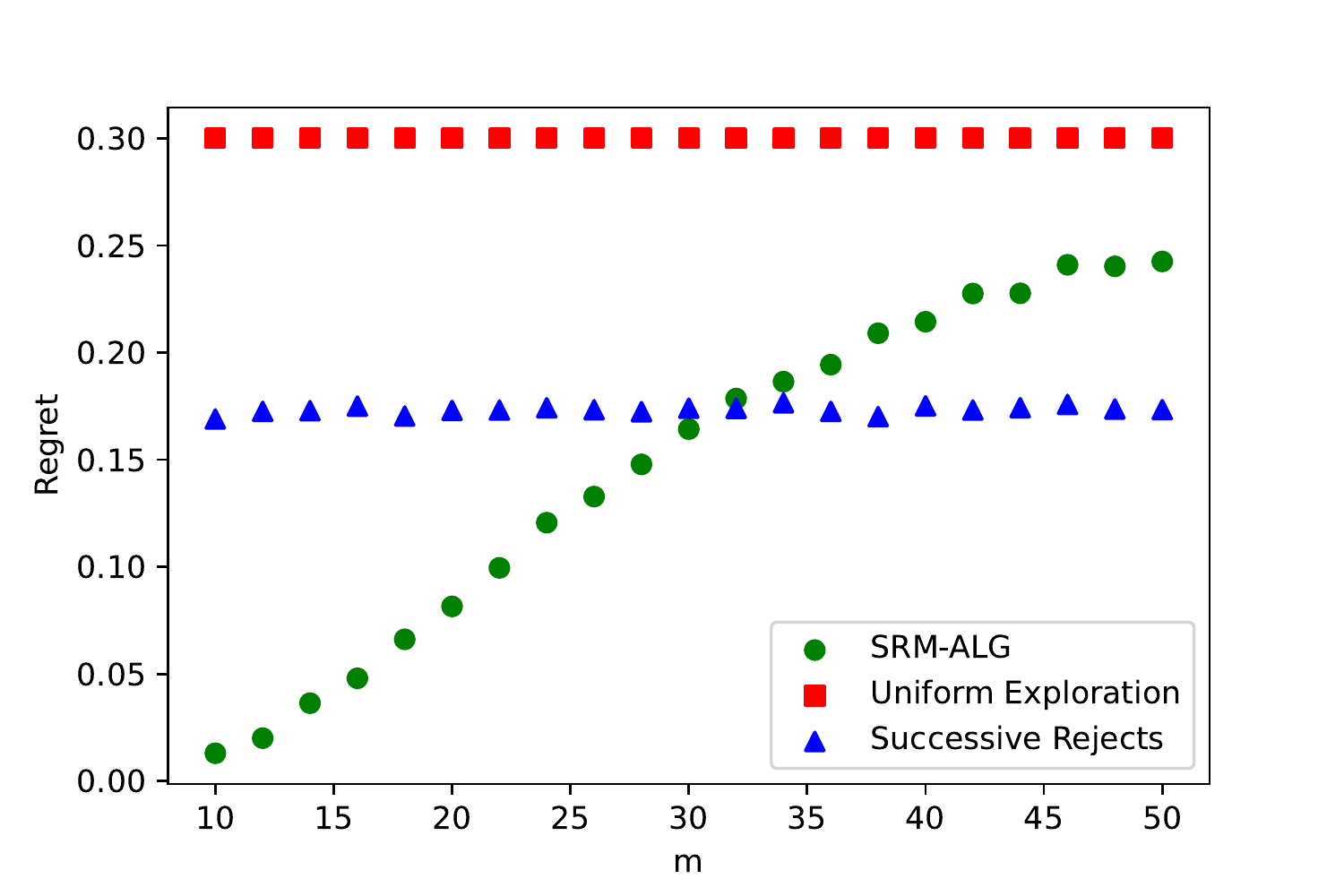}
        \caption[]{{\small Regret vs $m$, $N = 200$}} 	\label{fig:regret-vs-m-200}   
    \endminipage\hfill
    \minipage{0.5\textwidth}
      \includegraphics[width=\linewidth]{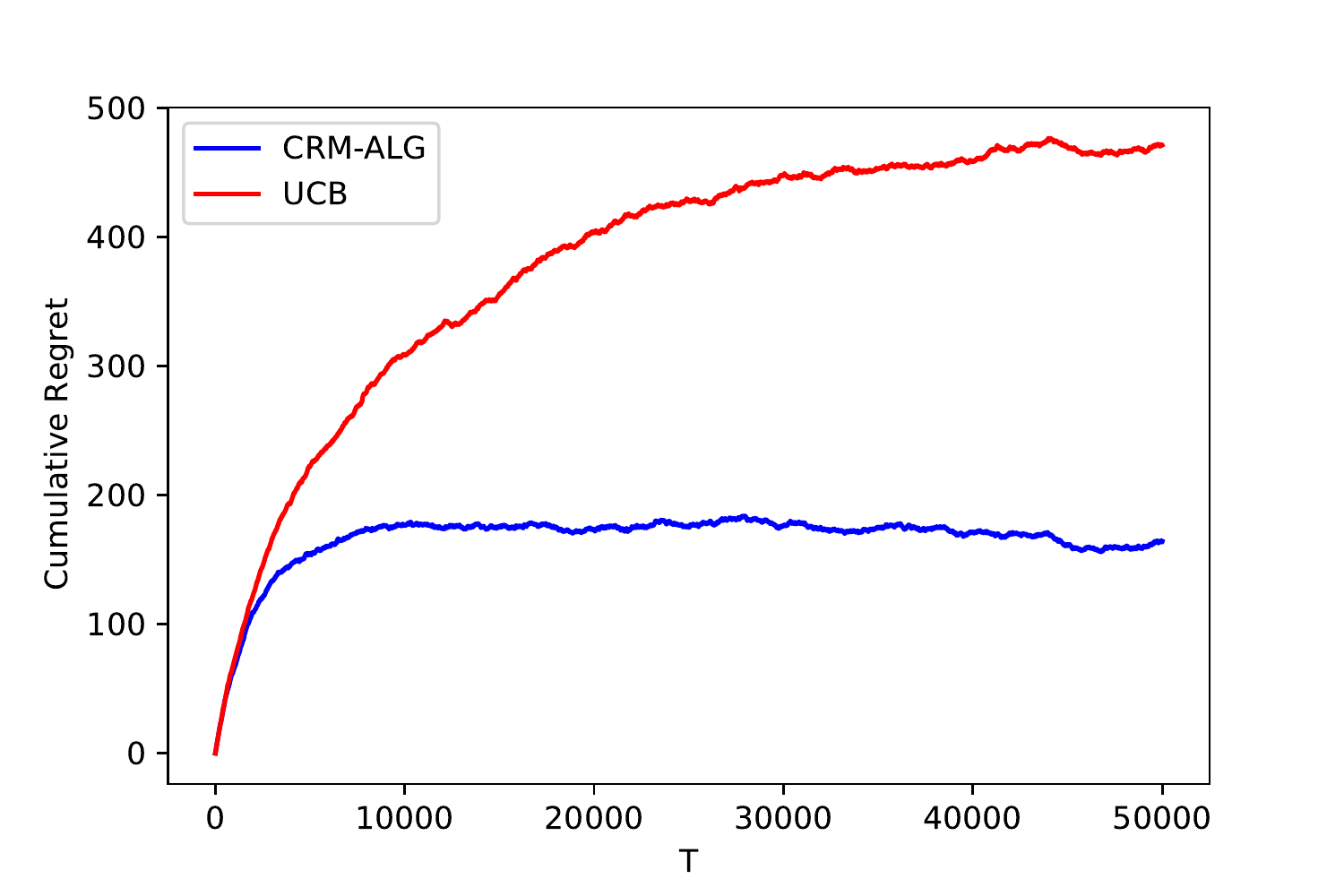}
        \caption[]{{\small Cumulative Regret vs Horizon}} 	\label{fig:cumulative-regret-vs-T}   
    \endminipage\hfill
\end{figure}

\textbf{Experiment $1$ (Simple Regret vs. T):} \VIN{This experiment compares the expected simple regret of \SRM\ with \UE\ and \SR\ as $T$ increases. We run the algorithms on 50 CBNs such that for every constructed CBN $C$, it has $100$ intervenable nodes and $m(\mathcal{C})=9$. The CBNs are constructed as follows: a) randomly generate $50$ DAGs on $101$ nodes $X_1, \ldots,X_{100}$ and $Y$, and let $X_1\prec \ldots \prec X_{100}\prec Y$ be the topological order in each such DAG,\ b) $\mathbf{Pa}(X_i)$ contains at most $2$ nodes chosen uniformly at random from $X_1, \ldots, X_{i-1}$, and $\mathbf{Pa}(Y)$ contains $X_i$ for all $i$,\ c) $\mathbb{P}(X_i \mid \mathbf{Pa}(X_i)) = 0.5$ for $i\in [91]$ and $\mathbb{P}(X_i|\mathbf{Pa}(X_i)) = 1/18$ for $i\in [92,100]$,\ d) uniformly at random choose a $j\in \{92,\ldots,100\}$ and set $P(Y|X_1,\ldots, X_j=1,\ldots, X_{100}) = 0.5 + \epsilon$ and  $P(Y|X_1,\ldots, X_j=0, \ldots, X_{100}) = 0.5 - \epsilon'$ where $\epsilon = 0.3$ and $\epsilon' = q\epsilon/(1-q)$ for $q=1/18$.
The choice of the conditional probability distributions (CPDs) in (c) ensures $m(\mathcal{C}) = 9$ for every CBN $\mathcal{C}$ that is generated. 
We note that the above strategy to generate CBNs is a generalization of of the one used in \cite{LattimoreLR16} to define parallel bandit instances with a fixed $m$.
For each of the $50$ random CBN, we run \SRM, \MAB, \SR\ for multiple values of the time horizon $T$ in $[500, 2500]$ and average the regret over $100$ independent runs. Finally, we calculate the mean regret over all the $50$ random CBNs and plot mean regret vs. $T$ in Fig. \ref{fig:regret-vs-T}. Since $m \ll N$, as seen in the Fig. \ref{fig:regret-vs-T}, \SRM\ has a much lower regret as compared to \UE\ and \SR\ which is in accordance with our regret bounds from Theorem \ref{theorem: UB-SR}.}\\

\textbf{Experiment $2$ (Simple Regret vs. m):} \VIN{This experiment compares the expected simple regret of \SRM\ with \UE\ and \SR\ for CBNs with different values of function $m$ from the set $S = \{10+2k : k\in [20]\}$. We fix $T=1600$, and randomly generate $50$ DAGs on $N+1$ nodes $X_1, \ldots,X_{N}$ and $Y$. For each generated DAG, $\mathcal{G}$ and $m \in S$, we use the same process as Experiment $1$ to set the CPDs of $\mathcal{G}$. We run \SRM, \MAB, \SR\ for $T$ time steps and average the regret over $100$ independent runs for each of the $50$ CBN, and repeat this experiment for $N=100$ and $N=200$. In Fig. \ref{fig:regret-vs-m-100}, we plot the mean regret over all the $50$ random CBNs vs. $m$ for $N=100$, and in Fig. \ref{fig:regret-vs-m-200} the same plot is provided for $N=200$. Our plots validate the $\sqrt{m}$ dependence of regret (for fixed $T$) in the case of \SRM. Notice that as $N$ increases, the regret of \SRM\ remains the same, that is, it only depends on $m$ (as shown in Theorem \ref{theorem: UB-SR}), whereas the regret of \MAB\ and \SR\ increases as expected. Hence, for larger values of $N$, \SRM\ is strictly better in terms of regret even for a wide range of values of $m$.}\\



\textbf{Experiment $3$ (Cumulative Regret vs. T):} This experiment compares the cumulative regret of \CRM\ with \UCB\ for CBN on four nodes $X_1, X_2, X_3$, and $Y$. Variable $X_1$ has no parents and is the only parent of $X_2, X_3$. Parents of $Y$ are $X_2, X_3$. We choose CPDs $\mathbb{P}(X_1=1) = 0.5$, $\mathbb{P}(X_2=1|X_1)$ and $\mathbb{P}(X_3=1|X_1)$ are equal to $0.75X_1 + 0.25(1-X_1)$ and $P(Y=1|X_2, X_3) = \mathds{1}_{X_2=X_3}$. For this instance, it is easy to see that $\mathbb{P}(Y=1|do(X_2=x)) = P(Y=1|do(X_3=x)) = 0.5$ for $x\in \{0,1\}$ and $P(Y=1|do()) = 5/8$, that is, the observational arm is the best arm. We run \CRM\ and \UCB\ for a sufficient time range $T$ and average the cumulative regrets over $30$ independent runs. Fig. \ref{fig:cumulative-regret-vs-T} demonstrates that while cumulative regret of \UCB\ increases, that of \CRM\ becomes constant for large enough $T$ as expected by result in Theorem \ref{theorem: UB-CRM}. We run another experiment with independently chosen random CBN instances without the guarantee of the observational arm being the best arm and for clarity of presentation discuss its results in Appendix \ref{secappendix:additional-experiments}.

\section{Conclusion}\label{sec: conclusion}
We proposed two algorithms \SRM\ and \CRM\ that minimize the simple and cumulative regrets respectively for general causal graphs with atomic interventions. \SRM\ also works when the input causal graph might have unobservable variables, whereas \CRM\ requires that all the variables in the input graph to be observable. We theoretically and empirically show that our proposed algorithms are better than standard \MAB\ that do not take into account the causal side-information. Further, we show that \SRM\ is almost optimal for any causal graph described as an $n$-ary tree. Importantly, our results on simple regret minimization subsume previous results proved assuming structural restrictions on the graphs.    \\

\VIN{ In \cite{NairPS21}, the observation intervention trade-off was studied, when observations are less-expensive compared to interventions. Since our algorithms \SRM\ and \CRM\ leverage the causal side information via observational arm pulls, they can be extended to such scenarios. An interesting future direction is to identify classes of causal graphs where better simple regret guarantee than \SRM\ can be attained. Another interesting question is to obtain simple regret guarantees in terms of the size of an intervention. In this work, we considered atomic intervention, which is in itself well-motivated, whereas \cite{YabeHSIKFK18} consider interventions of arbitrary sizes. Thus, it would be nice to determine how the regret depends on the size of the intervention.}  

\section*{Acknowledgement}
Vineet Nair is thankful to be supported by the European Union’s Horizon 2020 research and innovation program under grant agreement No 682203 -ERC-[ Inf-Speed-Tradeoff]. 

\bibliographystyle{plainnat}
\bibliography{references}

\begin{thebibliography}{23}
\providecommand{\natexlab}[1]{#1}
\providecommand{\url}[1]{\texttt{#1}}
\expandafter\ifx\csname urlstyle\endcsname\relax
  \providecommand{\doi}[1]{doi: #1}\else
  \providecommand{\doi}{doi: \begingroup \urlstyle{rm}\Url}\fi

\bibitem[Audibert et~al.(2010)Audibert, Bubeck, and Munos]{AudibertBM10}
Jean{-}Yves Audibert, S{\'{e}}bastien Bubeck, and R{\'{e}}mi Munos.
\newblock Best arm identification in multi-armed bandits.
\newblock In \emph{The 23rd Conference on Learning Theory {COLT}}, pages
  41--53. Omnipress, 2010.

\bibitem[Auer et~al.(1995)Auer, Cesa-Bianchi, Freund, and Schapire]{492488}
P.~Auer, N.~Cesa-Bianchi, Y.~Freund, and R.E. Schapire.
\newblock Gambling in a rigged casino: The adversarial multi-armed bandit
  problem.
\newblock In \emph{Proceedings of IEEE 36th Annual Foundations of Computer
  Science}, pages 322--331, 1995.
\newblock \doi{10.1109/SFCS.1995.492488}.

\bibitem[Auer et~al.(2002)Auer, Cesa{-}Bianchi, and Fischer]{AuerCF02}
Peter Auer, Nicol{\`{o}} Cesa{-}Bianchi, and Paul Fischer.
\newblock Finite-time analysis of the multiarmed bandit problem.
\newblock \emph{Mach. Learn.}, 47\penalty0 (2-3):\penalty0 235--256, 2002.

\bibitem[Bareinboim et~al.(2015)Bareinboim, Forney, and Pearl]{BareinboimFP15}
Elias Bareinboim, Andrew Forney, and Judea Pearl.
\newblock Bandits with unobserved confounders: {A} causal approach.
\newblock In \emph{Annual Conference on Neural Information Processing Systems,
  2015}, pages 1342--1350, 2015.

\bibitem[Bhattacharyya et~al.(2020)Bhattacharyya, Gayen, Kandasamy, Maran, and
  Vinodchandran]{BhattacharyyaGKMV20}
Arnab Bhattacharyya, Sutanu Gayen, Saravanan Kandasamy, Ashwin Maran, and N.~V.
  Vinodchandran.
\newblock Efficiently learning and sampling interventional distributions from
  observations.
\newblock \emph{CoRR}, abs/2002.04232, 2020.

\bibitem[Bottou et~al.(2013)Bottou, Peters, Qui\~{n}onero Candela, Charles,
  Chickering, Portugaly, Ray, Simard, and Snelson]{Bottou2013}
L\'{e}on Bottou, Jonas Peters, Joaquin Qui\~{n}onero Candela, Denis~X. Charles,
  D.~Max Chickering, Elon Portugaly, Dipankar Ray, Patrice Simard, and
  Ed~Snelson.
\newblock Counterfactual reasoning and learning systems: The example of
  computational advertising.
\newblock \emph{J. Mach. Learn. Res.}, 14\penalty0 (1):\penalty0 3207–3260,
  January 2013.
\newblock ISSN 1532-4435.

\bibitem[Greenewald et~al.(2019)Greenewald, Katz, Shanmugam, Magliacane,
  Kocaoglu, Adser{\`{a}}, and Bresler]{GreenewaldKSMKA19}
Kristjan~H. Greenewald, Dmitriy Katz, Karthikeyan Shanmugam, Sara Magliacane,
  Murat Kocaoglu, Enric~Boix Adser{\`{a}}, and Guy Bresler.
\newblock Sample efficient active learning of causal trees.
\newblock In \emph{Annual Conference on Neural Information Processing Systems,
  NeurIPS}, pages 14279--14289, 2019.

\bibitem[Haigh and Bessler(2004)]{Haigh2004}
Michael S. Haigh and David A. Bessler.
\newblock Causality and price discovery: An application of directed acyclic
  graphs.
\newblock \emph{The Journal of Business}, 77\penalty0 (4):\penalty0 1099--1121,
  2004.
\newblock ISSN 00219398, 15375374.

\bibitem[Lattimore et~al.(2016)Lattimore, Lattimore, and Reid]{LattimoreLR16}
Finnian Lattimore, Tor Lattimore, and Mark~D. Reid.
\newblock Causal bandits: Learning good interventions via causal inference.
\newblock In \emph{Annual Conference on Neural Information Processing Systems,
  2016}, pages 1181--1189, 2016.

\bibitem[Lee and Bareinboim(2018)]{LeeB18}
Sanghack Lee and Elias Bareinboim.
\newblock Structural causal bandits: Where to intervene?
\newblock In \emph{Annual Conference on Neural Information Processing Systems,
  2018}, pages 2573--2583, 2018.

\bibitem[Lee and Bareinboim(2019)]{LeeB19}
Sanghack Lee and Elias Bareinboim.
\newblock Structural causal bandits with non-manipulable variables.
\newblock In \emph{{AAAI} Conference on Artificial Intelligence, 2019}, pages
  4164--4172. {AAAI} Press, 2019.

\bibitem[Lu et~al.(2020)Lu, Meisami, Tewari, and Yan]{LU2020}
Yangyi Lu, Amirhossein Meisami, Ambuj Tewari, and William Yan.
\newblock Regret analysis of bandit problems with causal background knowledge.
\newblock In \emph{Conference on Uncertainty in Artificial Intelligence, 2020},
  pages 141--150. PMLR, 2020.

\bibitem[Lu et~al.(2021)Lu, Meisami, and Tewari]{lu2021causal}
Yangyi Lu, Amirhossein Meisami, and Ambuj Tewari.
\newblock Causal markov decision processes: Learning good interventions
  efficiently, 2021.

\bibitem[Nair et~al.(2021)Nair, Patil, and Sinha]{NairPS21}
Vineet Nair, Vishakha Patil, and Gaurav Sinha.
\newblock Budgeted and non-budgeted causal bandits.
\newblock In \emph{The 24th International Conference on Artificial Intelligence
  and Statistics, {AISTATS} 2021, April 13-15, 2021, Virtual Event}, volume 130
  of \emph{Proceedings of Machine Learning Research}, pages 2017--2025. {PMLR},
  2021.

\bibitem[Pearl(2000)]{Pearl00}
Judea Pearl.
\newblock \emph{Causality: Models, Reasoning, and Inference}.
\newblock Cambridge University Press, USA, 2000.
\newblock ISBN 0521773628.

\bibitem[Pearl(2009)]{PEARL2009}
Judea Pearl.
\newblock \emph{Causality}.
\newblock Cambridge university press, 2009.

\bibitem[Sen et~al.(2017{\natexlab{a}})Sen, Shanmugam, Dimakis, and
  Shakkottai]{SenSDS17}
Rajat Sen, Karthikeyan Shanmugam, Alexandros~G. Dimakis, and Sanjay Shakkottai.
\newblock Identifying best interventions through online importance sampling.
\newblock In \emph{International Conference on Machine Learning, 2017},
  volume~70 of \emph{Proceedings of Machine Learning Research}, pages
  3057--3066. {PMLR}, 2017{\natexlab{a}}.

\bibitem[Sen et~al.(2017{\natexlab{b}})Sen, Shanmugam, Kocaoglu, Dimakis, and
  Shakkottai]{SenSKDS17}
Rajat Sen, Karthikeyan Shanmugam, Murat Kocaoglu, Alexandros~G. Dimakis, and
  Sanjay Shakkottai.
\newblock Contextual bandits with latent confounders: An {NMF} approach.
\newblock In \emph{International Conference on Artificial Intelligence and
  Statistics, 2017}, volume~54 of \emph{Proceedings of Machine Learning
  Research}, pages 518--527. {PMLR}, 2017{\natexlab{b}}.

\bibitem[Slivkins(2019)]{Slivkins_Book}
Aleksandrs Slivkins.
\newblock Introduction to multi-armed bandits.
\newblock \emph{CoRR}, abs/1904.07272, 2019.
\newblock URL \url{http://arxiv.org/abs/1904.07272}.

\bibitem[Tian and Pearl(2002)]{TianP02}
Jin Tian and Judea Pearl.
\newblock A general identification condition for causal effects.
\newblock In \emph{National Conference on Artificial Intelligence and
  Conference on Innovative Applications of Artificial Intelligence, 2002},
  pages 567--573, 2002.

\bibitem[Tsybakov(2008)]{10.5555/1522486}
Alexandre~B. Tsybakov.
\newblock \emph{Introduction to Nonparametric Estimation}.
\newblock Springer Publishing Company, Incorporated, 1st edition, 2008.
\newblock ISBN 0387790519.

\bibitem[Velikova et~al.(2014)Velikova, {van Scheltinga}, Lucas, and
  Spaanderman]{Velikova2014}
Marina Velikova, Josien~Terwisscha {van Scheltinga}, Peter~J.F. Lucas, and Marc
  Spaanderman.
\newblock Exploiting causal functional relationships in bayesian network
  modelling for personalised healthcare.
\newblock \emph{International Journal of Approximate Reasoning}, 55\penalty0
  (1, Part 1):\penalty0 59--73, 2014.
\newblock ISSN 0888-613X.
\newblock \doi{https://doi.org/10.1016/j.ijar.2013.03.016}.
\newblock URL
  \url{https://www.sciencedirect.com/science/article/pii/S0888613X13000777}.
\newblock Applications of Bayesian Networks.

\bibitem[Yabe et~al.(2018)Yabe, Hatano, Sumita, Ito, Kakimura, Fukunaga, and
  Kawarabayashi]{YabeHSIKFK18}
Akihiro Yabe, Daisuke Hatano, Hanna Sumita, Shinji Ito, Naonori Kakimura,
  Takuro Fukunaga, and Ken{-}ichi Kawarabayashi.
\newblock Causal bandits with propagating inference.
\newblock In \emph{International Conference on Machine Learning, 2018},
  volume~80 of \emph{Proceedings of Machine Learning Research}, pages
  5508--5516. {PMLR}, 2018.

\end{thebibliography}

\newpage

\appendix
\section{Preliminary Lemmas}
We state couple of well-known and standard concentration bounds that would be used in the proving the theorems.
\begin{lemma}[Chernoff Bounds]{\label{lemma: Chernoff bound}}
Let $Z$ be any random variable. Then for any $t \geq 0$,
\begin{enumerate}
    \item $\mathbb{P}(Z \geq E[Z]+t) \leq \min_{\lambda \geq 0} E[e^{\lambda(Z - E[Z])}]e^{-\lambda t}$
    \item $\mathbb{P}(Z \leq E[Z]-t) \leq \min_{\lambda \geq 0} E[e^{\lambda(E[Z] - Z)}]e^{-\lambda t}$
\end{enumerate}
\end{lemma}
\vspace{0.1cm}
\begin{lemma}[Hoeffding's Lemma]{\label{lemma: Hoeffding's lemma}}
Let $Z$ be a bounded random variable with $Z \in [a, b]$. Then, $E[\exp(\lambda(Z-E[Z])] \leq \exp\big(\frac{\lambda^2(b-a)^2}{8}\big)$ for all $\lambda \in \mathbb{R}$.
\end{lemma}
\vspace{0.1cm}
\begin{lemma}[Chernoff-Hoeffeding inequality]\label{lemma: chernoff-hoeffding inequality}
Suppose $X_1, \ldots, X_T$ are independent random variables taking values in the interval $[0,1]$, and let $X = \sum_{t\in [T]} X_t$ and $\overline{X} = \frac{1}{T}(\sum_{t\in [T]} X_t)$. Then for any $\varepsilon \geq 0$ the following holds:
\begin{enumerate}
    \item $\mathbb{P}(\overline{X} - E[\overline{X}] \geq \varepsilon ) \leq
    e^{-2\varepsilon^2T}$
    \item $\mathbb{P}(\overline{X} - E[\overline{X}] \leq -\varepsilon ) \leq
    e^{-2\varepsilon^2T}$
\end{enumerate}
\end{lemma}
 \renewcommand{\labelenumi}{\alph{enumi})}

\section{Estimation of reward from observation}\label{secappendix: estimating rewards from observations}
In Algorithm \ref{alg: estimating rewards from observations} below we explain our strategy (derived from \cite{BhattacharyyaGKMV20}) for estimating the reward of the interventional arms $a_{i,x}$ using $T/2$ observational samples collected by playing the observational arm $a_0$. This is followed by further details on each of the steps involved.


\begin{algorithm}
\caption{Estimating Rewards from Observational Samples} \label{alg: estimating rewards from observations}
\begin{algorithmic}
\State INPUT: $\mathsf{His}$ containing the $T/2$ observational samples collected by playing arm $a_0$, and $\mathcal{G}$
\end{algorithmic}

\begin{algorithmic}[1]
\State For each $i\in [N]$, reduce the input ADMG $\mathcal{G}$ to ADMG $\mathcal{H}_i$ as outlined in Algorithm \ref{alg:graph-reduction}.
\State Next, for each $i\in [N]$ and $x\in \{0,1\}$, construct the Bayes net $D_{i,x}$ which simulates the causal effect of intervention $do(X_i=x)$ on the reduced graph $\mathcal{H}_i$.
\State Using Algorithm \ref{Dix-algorithm} on the input samples, learn the distributions of all $D_{i,x}$. Then, using learned $D_{i,x}$, generate samples to estimate marginal of $Y$ and return them as estimated rewards.
\end{algorithmic}
\end{algorithm}

\textbf{Step $1$ :} This step is executed using Algorithm \ref{alg:graph-reduction} based on the reduction algorithm from \cite{BhattacharyyaGKMV20}.
\begin{algorithm}
\caption{Reducing $\mathcal{G}$ to $\mathcal{H}_i$} 
\label{alg:graph-reduction}
\begin{algorithmic}
\State INPUT: ADMG $\mathcal{G}$ and index $i\in [N]$.
\end{algorithmic}
\begin{algorithmic}[1]
\State Let $\mathbf{W} = Y \cup X_i \cup \mathbf{Pa}^c(X_i)$, and $\mathcal{G}'_i$ be the graph obtained by considering $\mathbf{V}\backslash \mathbf{W}$ as hidden variables. Let $\mathbf{V}_i$ denote the nodes in $\mathcal{G}'_i$
\State \textbf{Projection Algorithm:} It reduces $\mathcal{G}'_i$ to $\mathcal{H}_i$ as follows:
  \begin{enumerate}
        \item Add all observable variables in $\mathcal{G}'_i$ to $\mathcal{H}_i$.
        \item For every pair of observable variable $V_j^i, V_k^i \in \mathbf{V}_i$, add a directed edge from $V_j^i$ to $V_k^i$ in $\mathcal{H}_i$, if $(a)$ there exists a directed edge from $V_j^i$ to $V_k^i$ in $\mathcal{G}_i'$, or if $(b)$ there exists a directed path from $V_j^i$ to $V_k^i$ in $\mathcal{G}'_i$ which contains only unobservable variables.
        \item For every pair of unobservable variable $V_j^i, V_k^i \in \mathbf{V}_i$, add a bi-directed edge between $V_j^i$ and $V_k^i$ in $\mathcal{H}_i$, if $(a)$ there exists an unobserved variable $U$ with two directed paths in $\mathcal{G}'_i$ going from $U$ to $V_j^i$ and $U$ to $V_k^i$ and containing only unobservable variables.
    \end{enumerate}
\State Return $\mathcal{H}_i$.
\end{algorithmic}
\end{algorithm}

\textbf{Step $2$ :} Construction of $D_{i,x}$ is done using the method described in Section $4.1$ of \cite{BhattacharyyaGKMV20}. Without loss of generality let ${\bf S_1}$ be the c-component containing $X_i$. To construct $D_{i,x}$, we start with $\mathcal{H}_i$. Then, for each $V \notin {\bf S_1}$ such that $X_i$ is in the set ${\bf Z_i}$ of ``effective parents'' (Section $4$,  \cite{BhattacharyyaGKMV20}) of $V$, we create a clone of $X_i$ and fix its value to $x$ (i.e. the clone has no parents). Then we remove all the outgoing edges from the original $X_i$. Note that, for any assignment $\bm{v}$ of all variables except $X_i$ in $\mathcal{H}_i$, the causal effect $\mathbb{P}_{\mathcal{H}_i}(\bm{v} | do(X_i = x)) = \sum_x \mathbb{P}_{D_{i,x}} (\bm{v}, X_i = x)$.


\textbf{Step $3$ :}
In this step, we learn the distribution of $D_{i,x}$ using the $T/2$ samples that were provided as input. Details are described in Algorithm \ref{Dix-algorithm}. Using this learned distribution, we get $O(T)$ samples and compute an empirical estimate $\widehat{\mu}_{i,x}$ of the reward $\mu_{i,x} = \mathbb{P}_{\mathcal{G}}(Y = 1 | do(X_i = x))$. This follows from the construction of $D_{i,x}$ in Step $2$ which implies,
\[
\mu_{i,x} = \mathbb{P}_\mathcal{G}(Y=1 | do(X_i = x)) = \mathbb{P}_{\mathcal{H}_i}(Y=1 | do(X_i = x)) = \sum_{x, \bm{v}^\prime} \mathbb{P}_{D_{i,x}} (Y=1, \bm{v}^\prime, X_i = x)
\]
where $\bm{v}^\prime$ is an assignment of nodes in $D_{i,x}$ other than $X_i$ and $Y$.

\begin{algorithm}
\caption{Learning $D_{i,x}$ and estimating $\mu_{i,x}$} \label{Dix-algorithm}
\begin{algorithmic}
\State INPUT: ADMG $\mathcal{H}_i$ and $x\in \{0,1\}$.
\end{algorithmic}
\begin{algorithmic}[1]
\For {every $V_j \in S_1$}
    \For {every assignment $V_j = v$ and $\mathbf{Z_j} = \mathbf{z}$ where $\mathbf{Z_j}$ are effective parents of $V_j$ in $\mathcal{H}_i$}
        \State $N_j \leftarrow$ the number of samples with $\mathbf{Z_j} = \mathbf{z}$
        \State $N_{j,v} \leftarrow$ the number of samples with $\mathbf{Z_j}=\mathbf{z}$ and $V_j = v$
        \State $\widehat{D}_{i,x}(V_j = v | \mathbf{Z_i} = \mathbf{z}) \leftarrow \frac{N_{j,v}+1}{N_j+2}$
    \EndFor
\EndFor
\For {every $V_j \in \mathbf{V_i} \backslash \mathbf{S_1}$}
    \For {every $V_j = v$ and $\mathbf{Z_j} \backslash {X_i} = \mathbf{z}$, where $\mathbf{Z_j}$ are effective parents of $V_j$ in $\mathcal{H}_i$}
        \If {$X \in \mathbf{Z_i}$}
            \State $N_{j} \leftarrow$ the number of samples with $\mathbf{Z_j} \backslash {X_i} = \mathbf{z}$ and $X_i = x$
            \State $N_{j,v} \leftarrow$ the number of samples with $V_j = v$, $\mathbf{Z_j} \backslash {X_i} = \mathbf{z}$ and $X_i = x$
            \If {$N_{j} \geq t$}
                \State $\widehat{D}_{i,x}(V_j = v | \mathbf{Z_i} = \mathbf{z}) \leftarrow \frac{N_{j,v}+1}{N_j+2}$
            \Else
                \State $\widehat{D}_{i,x}(V_j = v| \mathbf{Z_j} - \{X_i\} = \mathbf{z}, X_i = x) \leftarrow \frac{1}{2}$
            \EndIf
        \Else
            \State $N_{j} \leftarrow$ the number of samples with $\mathbf{Z_j} = \mathbf{z}$
            \State $N_{j,v} \leftarrow$ the number of samples with $V_j = v$ and $\mathbf{Z_j} = \mathbf{z}$
            \If {$N_{j} \geq t$}
                \State $\widehat{D}_{i,x}(V_j = v | \mathbf{Z_i} = \mathbf{z}) \leftarrow \frac{N_{j,v}+1}{N_j+2}$
            \Else
                \State $\widehat{D}_{i,x}(V_j = v | \mathbf{Z_j} = \mathbf{z}) \leftarrow \frac{1}{2}$
            \EndIf
        \EndIf
    \EndFor
\EndFor
\State Return $\widehat{D}_{i,x}$.
\end{algorithmic}
\end{algorithm}

\section{Proof of Theorem \ref{theorem: UB-SR}}\label{secappendix: proof of SRM}
\VIN{For the sake of analysis, we assume without loss of generality that $q_1, q_2, \ldots, q_{N}$ are arranged such that their corresponding c-component sizes $k_1, k_2, \ldots, k_{N}$ satisfy the following relation: $(q_1)^{k_1} \leq (q_2)^{k_2} \leq \ldots \leq (q_{N})^{k_{N}}$. Also, let $q = \min_{i\{q_i > 0\}} q_i$ (if $q_i = 0$ for all $i \in [N]$ then $q = \frac{1}{N+1}$), $k = \max_i k_i$, and $p^{i,x}_{\mathbf{z}} = \mathbb{P}(X_i = x, \mathbf{Pa}^c(X_i) = \mathbf{z})$. We remark that $p^{i,x}_{\mathbf{z}}$ is different from $p^{i,x}_{\mathbf{z}}$ used in Section \ref{sec: cumulative regret} to denote $\mathbb{P}\{X_i=x, \mathbf{Pa}(X_i) =\mathbf{z}\}$; note that $\mathbf{Pa}(X_i) \subseteq \mathbf{Pa}^c(X_i)$. Finally, let $Z_i$ be the size of the domain from which $\mathbf{Pa}^c(X_i)$ takes values, and note that $Z_i \leq 2^{k_id+k_i}$ and let $Z = \max_i Z_i$.}

We begin by proving Lemmas \ref{q-estimate-lemma}, \ref{m-estimate-lemma}, and \ref{mu-estimation-lemma} which would be used to prove Theorem \ref{theorem: UB-SR}.  The following lemma bounds the probability of making a bad estimate of $q_i$ for any $i \in [N]$, at the end of $T/2$ rounds.

\VIN{\begin{lemma}
\label{q-estimate-lemma}
Let $F =\mathds{1}\{\text{At the end of } T/2 \textit{ rounds, there exists } i \textit{ such that } |\widehat{q}_i - q_i| \geq \frac{1}{4}(1-2^{-1/k})q \}$. Then $\mathbb{P}(F = 1) \leq 4NZe^{-\frac{1}{16}(1-2^{-1/k})^2 q^2 T}$.
\end{lemma}}
\begin{proof}
Let $F_{i,x} = \mathds{1}\{\textit{At the end of } T/2 \textit{ rounds there exists } \mathbf{z} \textit{ such that } |\widehat{p}_{\mathbf{z}}^{i,x} - p_{\mathbf{z}}^{i,x}| \geq \frac{1}{4}(1-2^{-1/k})q\}$. From Lemma \ref{lemma: chernoff-hoeffding inequality}, it follows that,
\begin{align*}
    \mathbb{P}(|\widehat{p}_{\mathbf{z}}^{\ i,x} - p_{\mathbf{z}}^{i,x}| \geq \frac{1}{4}(1-2^{-1/k})q) \leq 2e^{-2\frac{1}{16}(1-2^{-1/k})^2 q^2 \frac{T}{2}}
\end{align*}
By union bound,
\begin{align*}
    \mathbb{P}(F_{i,x} = 1) \leq 2Z_ie^{-\frac{1}{16}(1-2^{-1/k})^2 q^2 T}
\end{align*}
By definition $q_i = \min_{x, \mathbf{z}} p_{\mathbf{z}}^{i,x}$ and $\widehat{q}_{i} = \min_{x, \mathbf{z}} \widehat{p}_{\mathbf{z}}^{\ i,x}$. Hence,
\begin{align*}
    \mathbb{P}(|\widehat{q}_{i} - q_{i}| \geq \frac{1}{4}(1-2^{-1/k})q) \leq 2P(F_{i,x} = 1) \leq 4Z_ie^{-\frac{1}{16}(1-2^{-1/k})^2 q^2 T}
\end{align*}
Taking union bound, we get $\mathbb{P}(F=1) \leq 4NZe^{-\frac{1}{16}(1-2^{-1/k})^2 q^2 T}$.
\end{proof}

\VIN{The next lemma shows that with high probability the estimate of $m$ at Step 6 of \SRM\ is good.}
\VIN{
\begin{lemma}
\label{m-estimate-lemma}
Let $F$ be as defined in Lemma \ref{q-estimate-lemma} and let $J = \mathds{1}\{\text{At the end of $T/2$ rounds the following holds } \hat{m} \leq 2m $. Then $F = 0$ implies $J = 1$, and in particular, $\mathbb{P}(J = 1) \geq 1-4NZe^{-\frac{1}{16}(1-2^{-1/k})^2 q^2 T}$.
\end{lemma}
}
\begin{proof}
Note that if $q_i = 0$ for all $i \in [N]$, then our proposition is trivially true. $F = 0$ implies after $T/2$ rounds for all $i \in [N]$, $|\widehat{q}_{i} - q_{i}| \leq \frac{1}{4}(1-2^{-1/k})q$. Now from definition of $m$ we know that there is an $l \leq m$ such that for $i > l$, $(q_i)^{k_i} \geq (\frac{1}{m})$. Hence, for $i > l$, since $q \leq q_i$ by definition \\
\begin{align*}
(\widehat{q}_i)^{k_i} \geq (q_i - \frac{1}{4}(1-2^{-1/k})q)^{k_i} \geq (q_i - (1-2^{-1/k})q_i)^{k_i} \geq \frac{1}{2^{k_i/k} m} \geq \frac{1}{2m}
\end{align*}

Since, $l \leq m$, we have $|\{j | \widehat{q}_j^{k_j} < \frac{1}{2m}\}| \leq 2m$. This implies $\widehat{m} \leq 2m$.

\end{proof}


The next lemma provides the confidence bound on the estimate of $\mu_{i,x}$ computed by Algorithm \ref{alg: estimating rewards from observations} for each $i,x$ .
\VIN{
\begin{lemma}
\label{mu-estimation-lemma}
For an action $a_{i,x} \in \mathcal{A}$, at the end of $T/2$ rounds $\mathbb{P}(|\widehat{\mu}_{i,x} - \mu_{i,x}| > \epsilon) 
\leq \exp{\big({-\epsilon^2 \frac{q_{i}^{k_i} T}{K_{\mathcal{G}}}}\big)}$, where $K_{\mathcal{G}} \geq 1$ is a constant dependent on the structure of $\mathcal{G}$ but independent of $\mathbb{P}$.
\end{lemma}
}
\begin{proof}
Using \textbf{Theorem 2.5} and \textbf{Theorem A.1} in \cite{BhattacharyyaGKMV20}, it can be inferred that the learner can estimate $\hat{\mu}_{i,x}$, such that $|\hat{\mu}_{i,x} - \mu_{i,x}| \leq \epsilon$, with probability $1-\delta_i$, using $O\big(2^{2u_i^2} \log 2^{2u_i^2} \log \frac{1}{\delta_i} /(q_{i}^{k_i} \epsilon^2)\big)$ samples, where $u_i = 1 + k_i(d+1)$. Hence using samples $T = K'\frac{2^{2.2u_i^2}}{q_{i}^{k_i} \epsilon^2} \log \frac{1}{\delta_i}$, where $K'$ is a constant independent of the problem instance, we get, $P(|\hat{\mu}_{i,x} - \mu_{i,x}| \leq \epsilon) \geq 1-\delta_i$. Writing $\delta_i$ in terms of $T$ and $\epsilon$, and using $K_{\mathcal{G}} = \max \{1, K' 2^{2.2u_i^2}\}$,
\begin{align*}
    \mathbb{P}(|\hat{\mu}_{i,x} - \mu_{i,x}| > \epsilon) \leq \exp{\bigg(-\frac{T}{K'} \frac{q_{i}^{k_i} \epsilon^2}{2^{2.2u_i^2}}\bigg)} \leq \exp{\bigg({-\epsilon^2 \frac{q_{i}^{k_i} T}{K_{\mathcal{G}}}}\bigg)}
\end{align*}
Also by \ref{lemma: chernoff-hoeffding inequality}, for $a_0$, by,
\begin{align*}
    \mathbb{P}(|\hat{\mu}_0 - \mu_0| \geq \epsilon) \leq \exp{\bigg(-2 \epsilon^2 \frac{T}{2}\bigg)}~.
\end{align*}
\end{proof}
Now we are ready to prove the theorem using the above Lemmas, and let $K = 2^{k-1}K_{\mathcal{G}}$. \AUR{Let $L_1 = \min_{t\in \mathbb{N}}(4NZe^{-\frac{1}{16}(1-2^{-1/k})^2q^2t} \leq \sqrt{\frac{144Km}{t}\log\frac{Nt}{m}})$ and $L_2 = \min_{t \in \mathbb{N}} \frac{6}{N^3}(\frac{m}{t})^4 \leq \sqrt{\frac{16Km}{t}\log\frac{Nt}{m}}$ and we assume throughout the proof that $T \geq \max\{L_1, L_2\}$}.
Consider $a_{i,x} \in \mathcal{Q}$. By Lemma \ref{lemma: chernoff-hoeffding inequality}, and Lemma \ref{m-estimate-lemma},
\begin{align*}
    \mathbb{P}\{|\widehat{\mu}_{i,x} - \mu_{i,x}| \geq \epsilon | F = 0\} \leq 2\exp{\bigg({-\epsilon^2\frac{2T}{4\widehat{m}}}\bigg)} \leq 2\exp{\bigg({-\epsilon^2\frac{T}{4m}}\bigg)} \leq 2\exp{\bigg({-\epsilon^2\frac{T}{4K m}}\bigg)} 
\end{align*}
If $a_{i,x} \notin \mathcal{Q}$, and $q_i^{k_i} \geq \frac{1}{m}$, then given $F = 0$ we get,
\begin{align*}
    &\mathbb{P}\{|\widehat{\mu}_{i,x} - \mu_{i,x}| > \epsilon | F = 0\} \leq \exp{\bigg({-\epsilon^2 \frac{q_{i}^{k_i} T}{K_{\mathcal{G}}}}\bigg)} \leq  \exp{\bigg({-\epsilon^2 \frac{T}{4K m}}\bigg)}
\end{align*}

If $a_{i,x} \notin \mathcal{Q}$, and $q_i^{k_i} < \frac{1}{m}$, then given $F = 0$ from Lemma \ref{q-estimate-lemma}, $q_i^{k_i} \geq (\widehat{q}_i - \frac{1}{4}(1-2^{-1/k})q)^{k_i} \geq ((\frac{1}{\widehat{m}})^{1/k_i} - \frac{1}{4}(\frac{1}{m})^{1/k_i}))^{k_i} \geq ((\frac{1}{2m})^{1/k_i} - \frac{1}{4}(\frac{1}{m})^{1/k_i}))^{k_i} \geq \frac{1}{2^{k+1}m}$ we get,
\begin{align*}
    &\mathbb{P}\{|\widehat{\mu}_{i,x} - \mu_{i,x}| > \epsilon | F = 0\} \leq \exp{\bigg({-\epsilon^2 \frac{q_{i}^{k_i} T}{K_{\mathcal{G}}}}\bigg)} \leq \exp{\bigg({-\epsilon^2 \frac{T}{2^{k+1}K_{\mathcal{G}}m}}\bigg)} \leq \exp{\bigg({-\epsilon^2 \frac{T}{4Km}}\bigg)}
\end{align*}
\begin{align*}
    \mathbb{P}\{ \textit{There exists an action $a$ such that } |\hat{\mu}_a - \mu_a| > \epsilon | F = 0 \} &\leq (4N + 2)\exp{\bigg({-\epsilon^2 \frac{T}{4K m}}\bigg)} \\
    &\leq 6N\exp{\bigg({-\epsilon^2 \frac{T}{4K m}}\bigg)}
\end{align*}
Substituting $\epsilon = \sqrt{\frac{16K m}{T}\log\frac{NT}{m}}$, we get,
\begin{align*}
E[r_T | F=0] \leq 2 \sqrt{\frac{16K m}{T}\log{\frac{NT}{m}}} + \frac{6}{N^3}\bigg(\frac{ m}{T}\bigg)^4 \leq \sqrt{\frac{144K m}{T}\log{\frac{NT}{m}}}
\end{align*}
Finally, the expected simple regret of Algorithm \ref{SR-algorithm} is as follows:
\begin{align*}
E[r_T] &= E[r_T | F = 0 ]\mathbb{P}(F = 0) + E[r_T | F = 1 ]\mathbb{P}(F = 1) \nonumber \\
&\leq E[r_T | F = 0 ] + \mathbb{P}(F = 1) \nonumber \\
&\leq \sqrt{\frac{144K m}{T}\log{\frac{NT}{m}}} + 4NZe^{-\frac{1}{16}(1-2^{-1/k})^2 q^2 T}
\end{align*}
Since $T \geq \max(L_1, L_2)$ the simple regret is $\mathcal{O}\bigg(\sqrt{\frac{m}{T} \log{\frac{NT}{m}}}\bigg)$.

\section{Proof of Theorem \ref{theorem: LB-Tree}}\label{secappendix: proof of lower bound for tree}


\VIN{Throughout this proof we assume the following terminology: a) a node is a root node if it has not parents, b) a node is a leaf node if it has no children. Consider an n-ary tree $\mathcal{T} \in \mathsf{T}$ on $N$ intervenable nodes. Note that since $\mathcal{T}$ is a tree, each node $X_i$ for $i\in [N]$ has at most one parent. In addition $\mathcal{T}$ has one special node $Y$, called the outcome. There is a directed from every leaf node in $\mathcal{T}$ to $Y$, and let $L_{\mathcal{T}}$ be the set of all leaf nodes. We use $\mathbf{V}$ to denote the set of nodes in $\mathcal{T}$, that is, $\mathbf{V} = \{X_1, \ldots, X_N, Y\}$. Without loss of generality, we assume that $X_1, \ldots, X_N$ is in the reverse topological order, that is, $X_1$ is a leaf node, $X_N$ is a root node, $X_{N-1}$ is either a root node or a child of $X_N$, and so on. Let $\mathcal{T}_M$ be the sub-graph of $\mathcal{T}$ defined by the nodes $X_1, \ldots, X_M$. An edge belongs to $\mathcal{T}_M$ if both its endpoints belong to $\{X_1, \ldots, X_M\}$. Further, let $h$ be the maximum number of nodes in a (directed) path from a root node to $Y$. Now we define distributions $\mathbb{P}_{0}, \ldots , \mathbb{P}_{M}$ all compatible with $\mathcal{T}$ such that the optimal arm in the CBN $\mathcal{C}_i = (\mathcal{T}, \mathbb{P}_i)$ is $a_{i,1}$ for $i\in [M]$, and for $\mathcal{C}_0 = (\mathcal{T}, \mathbb{P}_0)$ every arm is an optimal arm.}

\textbf{Defining} $\mathbb{P}_0$: For $X_i$ not belonging to $\mathcal{T}_M$ let $\mathbb{P}_0(X_i = 1) = 0.5$, and for $X_i$ belonging to $\mathcal{T}_M$ and for an appropriately chosen $\alpha$ let 
\begin{align*}
&\mathbb{P}_0(X_i = 1) = \alpha ~~~&\text{ If }X_i \text{ is a root node,} \\
&\mathbb{P}_0(X_i = 1 | \mathbf{Pa}(X_i) = 0) = \alpha ~~~&\text{ If }X_i \text{ is not a root node,} \\
&\mathbb{P}_0(X_i = 1 | \mathbf{Pa}(X_i) = 1) = 1-\alpha ~~~&\text{ If }X_i \text{ is not a root node,} \\
&\mathbb{P}_0(Y=1| . ) = 0.5 ~~~~ \mathbb{P}_0(Y=0 | . ) = 0.5 
\end{align*}
The value of $\alpha$ is appropriately chosen later to achieve the desired lower bound. Note that in the above equations if $X_i$ is not a root node then $\mathbf{Pa}(X_i)$ is a singleton set. Also, $\mathbb{P}_0(Y=1| . )$ denotes the probability of $Y=1$ conditioned on any value of its parents. Next, we define $\mathbb{P}_i$ for $i\in [N]$.

\textbf{Defining} $\mathbb{P}_i$: Let $L_i$ be the set of leaf nodes that are reachable from $X_i$, that is there is a directed path from $X_i$ to every leaf node in $L_i$. Note that if $X_i$ is a leaf then $L_i = \{X_i\}$. We use $L_i = \mathbf{1}$ and $L_i = \mathbf{0}$ to denote all nodes in $L_i$ evaluated to $1$ and $0$ respectively. Also, let $L_{\mathcal{T}}^M$ be the set of all leaves in $\mathcal{T}_M$ and $L'_i = L_{\mathcal{T}}^M \setminus L_i$. Then
\begin{align*}
\mathbb{P}_i(Y | L_i = \mathbf{1}, L'_i = \mathbf{0}) = 0.5 + \epsilon~.
\end{align*}
The value of $\epsilon$ is appropriately chosen later to achieve the desired lower bound. The distributions of $X_i$ given its parents corresponding to $\mathbb{P}_i$ is the same as those defined for $\mathbb{P}_0$. \\

We set $\alpha = \min\{(2h|L_\mathcal{T}| + 2^{h+1})^{-1}, (2^{h}|L_\mathcal{T}|M)^{-1}\}$ and hence $\alpha < \frac{1}{M}$. Using this it is easy to see that $m(\mathcal{C}_i) = M$ for $i\in [0,M]$, and $M > 4$. Additionally, in $\mathcal{C}_i$ arm $a_{i,1}$ is the optimal arm for $i\in [1,M]$ and the reward for every arm in $\mathcal{C}_0$ is $0.5$.
We will denote $a^*$ as the optimal arm for every $\mathcal{C}_i$, and note that $a^* = a_{i,1}$ for $\mathcal{C}_i$, where $i\in [M]$. First, in Lemma \ref{lemma: lower bound on regret}, we lower bound the regret of returning a sub-optimal arm in $\mathcal{C}_i$ at the end of $T$ rounds. Further, in Lemma \ref{lemma: lower bound on the probability of choosing a sub-optimal arm}, we show that any algorithm would have a non-trivial probability of returning a sub-optimal arm in at least one of the constructed CBNs. Finally, we would use Lemmas \ref{lemma: lower bound on regret} and \ref{lemma: lower bound on the probability of choosing a sub-optimal arm} to lower bound the expected regret of any algorithm. 
\VIN{Let $\text{rew}_i(a_{j, x})$ denote the expected reward of action $do(X_j=x)$ under the distribution $\mathbb{P}_i$. We deviate from the usual notation of $\mu$ in this case, because the reward now depends on the arm and the corresponding distribution.} We require the following sets in Lemmas \ref{lemma: lower bound on regret} and \ref{lemma: lower bound on the probability of choosing a sub-optimal arm}: $V_1 = L_i \setminus L_j$, $V_2 = L_i \cap L_j$, $V_3 = L_j \setminus L_i$, $V_4 = L_{\mathcal{T}}^M \setminus (L_i \cup L_j)$, and $V_5 = V \setminus L_{\mathcal{T}}^M$. 

\begin{lemma}\label{lemma: lower bound on regret}
For every $i \in [1,M]$, $j \in [1,N]$, $x\in \{0,1\}$, and $(j,x) \neq (i,1)$ the following holds: $\text{rew}_i(a_{i,1}) - \text{rew}_i(a_{j,x}) \geq 0.5\epsilon$.
\end{lemma}
\begin{proof}
For any $i,j \in [M]$, we have
\begin{align}
\text{rew}_i(a_{i,1}) &= 0.5 + \mathbb{P}_i(V_4 = \mathbf{0}, V_1=\mathbf{1}, V_2=\mathbf{1}, V_3=\mathbf{0} \mid do(X_i =1))(\epsilon) \label{eqn: rew for i} \\
\text{rew}_i(a_{j,1}) &= 0.5 + \mathbb{P}_i(V_4 = \mathbf{0}, V_1=\mathbf{1}, V_2=\mathbf{1}, V_3=\mathbf{0}\mid do(X_j =1))(\epsilon) \label{eqn: rew for j}
\end{align}
Subtracting Equation \ref{eqn: rew for j} from Equation \ref{eqn: rew for i} we have
\begin{align*}
&\text{rew}_i(a_{i,1}) - \text{rew}_i(a_{j,1}) \nonumber\\
&= \mathbb{P}_i(V_4 = \mathbf{0}) \big[\mathbb{P}_i(V_1=\mathbf{1}, V_2=\mathbf{1}, V_3=\mathbf{0} \mid do(X_i = 1)) - \mathbb{P}_i(V_1=\mathbf{1}, V_2=\mathbf{1}, V_3=\mathbf{0} \mid do(X_j = 1))\big] \epsilon \nonumber\\
&= \mathbb{P}_i(V_4 = \mathbf{0}) \big[\mathbb{P}_i(V_3 = \mathbf{0}) \mathbb{P}_i(V_1=\mathbf{1}, V_2=\mathbf{1} \mid do(X_i = 1)) - \mathbb{P}_i(V_1=\mathbf{1}) P(V_2=\mathbf{1}, V_3=\mathbf{0} \mid do(X_j = 1))\big] \epsilon \nonumber\\
&\underset{(i)}{\geq} (1-\alpha)^{h|V_4|}\big[(1-\alpha)^{h(|L_i|+|V_3|)} - (2^h\alpha)\big] \epsilon \nonumber \\
&\geq ((1-\alpha)^{h|L_\mathcal{T}|} - 2^h\alpha)\epsilon \nonumber \\
&\geq ((1-h|L_\mathcal{T}|\alpha) - 2^h\alpha)\epsilon \nonumber \\
&\geq 0.5\epsilon
\end{align*}
(i) in the above equations follows from the definitions of $h$ and $\mathbb{P}_i$.
Similarly, it can be shown that $\text{rew}_i(a_{i,1}) - \text{rew}_i(a_{j,0}) \geq 0.5 \epsilon$ for $j \in [N]$, and $\text{rew}_i(a_{i,1}) - \text{rew}_i(a_{j,1}) \geq 0.5 \epsilon$ for $j \in [M+1,N]$. Also $\text{rew}_i(a_{i,1}) - \text{rew}_i(a_0) \geq 0.5 \epsilon$.
\end{proof}
Let \texttt{ALG} be an algorithm that outputs arm $a_T$ at the end of $T$ rounds. We choose $\epsilon = \min \{\frac{1}{4}, \sqrt{\frac{M}{18T}}\}$. Note that corresponding to every $\mathcal{C}_i$ for $i\in [0,M]$, \texttt{ALG} and $\mathbb{P}_i$ together define a probability measure on all the sampled values of the nodes of $\mathcal{T}$ over $T$ rounds. Denote $\mathbb{D}_i$ as this measure and $E_i$ as the expectation over $\mathbb{D}_i$ for $i \in [0,M]$. Let $\mathcal{G}_{t}$ be the sampled values of the nodes of $\mathcal{T}$ at time $t$ and let $\mathbf{G}_{t} = \{\mathcal{G}_{1}, \ldots, \mathcal{G}_{t}\}$. Also, for $i \in [0,M]$ let $\mathbb{D}_i(.|\mathbf{G}_{t-1}) = \mathbb{P}_i^t(.)$; here $\mathbb{D}_i(.|\mathbf{G}_{t-1})$ denotes the probability of the sampled values of the nodes of $\mathcal{G}$ conditioned on its history till time $t-1$. Observe that conditioned on history $\mathbf{G}_{t-1}$, $\texttt{ALG}$ determines an arm, say $a_t$, to pull  at time $t$ (either deterministically or in a randomized way), and for $j,j'\in [1,N]$ if $a_t = a_{j,x}$ then $\mathbb{P}_i^t(X_{j'} = x | do(X_j)=x) = \mathbb{P}_i(X_{j'}=x| do(X_j=x))$.

\begin{lemma}\label{lemma: lower bound on the probability of choosing a sub-optimal arm}
For any algorithm \texttt{ALG} there exists an $i \in [M]$ such that 
$\mathbb{D}_i\{a_T \neq a_{i,1}\} \geq \frac{\frac{M}{4e} - 1}{M}$.   
\end{lemma}
\begin{proof}
We use $KL(\mathbb{D}_0, \mathbb{D}_i)$ to denote the KL divergence between $\mathbb{D}_0$ and $\mathbb{D}_i$ for any $i\in [M]$. Let $N^{(i,1)}_T$ be the number of times  \texttt{ALG} plays the arm $a_{i,1}$ at the end of $T$ rounds. Also, let $\mathcal{B} = \{ a_{i,1} \mid  i \leq M \textit{ and } E_0[N^{(i,1)}_T] \leq 2T/M \}$. Observe that $|\mathcal{B}| \geq M/2$, as otherwise the sum of the expected number of arm pulls of arms not in $\mathcal{B}$ would be greater than $T$. First, using Lemma 2.6 from \cite{10.5555/1522486}, we have, 
\begin{equation*}
\mathbb{D}_0(a_T = a_{i,1}) + \mathbb{D}_i(a_T \neq a_{i,1}) \geq 
\frac{1}{2}\cdot  \exp{(-KL(\mathbb{D}_0, \mathbb{D}_i))} 
\end{equation*}
Rearranging and summing the above equation over arms in $\mathcal{B}$, and observing that $\sum_{a_{i,1} \in \mathcal{B}} \mathbb{D}_0(a_T = a_{i,1}) \leq 1$ we have
\begin{equation}\label{equation: lower bound in terms of KL Divergence}
\sum_{a_{i,1} \in \mathcal{B}} \mathbb{D}_{i}\{a_T \neq a_{i,1}\} \geq \frac{1}{2}\cdot \sum_{a_{i,1} \in \mathcal{B}} \exp(-KL(\mathbb{D}_0, \mathbb{D}_{i}))-1 
\end{equation} 
Now we bound $\exp(-KL(\mathbb{D}_0, \mathbb{D}_{i}))$ for every $i$ such that $a_{i,1} \in \mathcal{B}$.
Using the chain rule for product distributions (see \cite{492488} and Chapter 2 in \cite{Slivkins_Book}) the KL divergence of $\mathbb{D}_0$ and $\mathbb{D}_i$ for any $i \in [M]$ can be written as 
\begin{align}
KL(\mathbb{D}_0, \mathbb{D}_i) = \sum_{t=1}^T KL(\mathbb{D}_0(\mathcal{G}_{t} | \mathbf{G}_{t-1}), \mathbb{D}_i(\mathcal{G}_{t} | \mathbf{G}_{t-1}) = \sum_{t=1}^T KL(\mathbb{P}_0^t(\mathcal{G}_t), \mathbb{P}_i^t(\mathcal{G}_t)) \label{kl-divergence-summation}
\end{align}
Each term on the right hand side of the above summation can be computed as follows:
\begin{align}
KL(\mathbb{P}_0^t, \mathbb{P}_i^t) &= \sum_{\mathbf{v}} \mathbb{P}_0^t(V = \mathbf{v}) \log \frac{\mathbb{P}_0^t(\mathbf{V} = \mathbf{v})}{\mathbb{P}_i^t(\mathbf{V}= \mathbf{v})} \nonumber \\
    &\underset{(i)}{=} \sum_{x,\mathbf{v_5}} \mathbb{P}_0^t(Y=x, L_i = \mathbf{1}, L'_i = \mathbf{0}, V_5 = \mathbf{v_5}) \log \frac{\mathbb{P}_0^t(Y=x | L_i = \mathbf{1}, L'_i = \mathbf{0}, V_5 = \mathbf{v_5})}{\mathbb{P}_i^t(Y=x | L_i = \mathbf{1}, L'_i = \mathbf{0}, V_5 = \mathbf{v_5})} \nonumber \\
    &\underset{(ii)}{=} 0.5 \cdot \mathbb{P}_0^t(L_i = \mathbf{1}, L'_i = \mathbf{0}) \Big[ \log\frac{0.5}{0.5 + \epsilon} + \log\frac{0.5}{0.5 - \epsilon} \Big] \nonumber \\
    &\underset{(iii)}{\leq} 0.5 \Big(\mathbb{P}_0^t\{do(X_i=1)\} + 2^h|L_\mathcal{T}|\alpha \Big) \log\frac{0.25}{0.25 - \epsilon^2} \nonumber \\
    &= -0.5 \Big(\mathbb{P}_0^t\{do(X_i=1)\} + 2^h|L_\mathcal{T}|\alpha \Big) \log(1-4\epsilon^2) \nonumber \\
    &= 0.5 \Big(\mathbb{P}_0^t\{do(X_i=1)\} + 2^h|L_\mathcal{T}|\alpha \Big) \Big(4\epsilon^2 + \frac{(4\epsilon^2)^2}{2} + \frac{(4\epsilon^2)^3}{3} + \dots \Big) \nonumber \\
    &\leq 6 \Big(\mathbb{P}_0^t\{do(X_i=1)\} + 2^h|L_\mathcal{T}|\alpha \Big) \epsilon^2~. \label{equation: individual term KL divergence}
\end{align}
In the above equations: (i) follows by observing that for every other evaluation of $\mathbf{V}$ the distributions $\mathbb{P}_0^t$ and $\mathbb{P}_i^{t}$ are same hence the corresponding terms in KL divergence amount to zero, (ii) follows from the definitions of $\mathbb{P}_0^t$ and $\mathbb{P}_i^t$, and (iii) follows by observing that 
$$\mathbb{P}_0^t(L_i = \mathbf{1}, L'_i = \mathbf{0}) \leq \mathbb{P}_0^t\{do(X_i=1)\} + 2^h|L_\mathcal{T}|\alpha ~.$$
Using Equations \ref{kl-divergence-summation} and \ref{equation: individual term KL divergence}, we have
for every $a_{i,1} \in \mathcal{B}$,
\begin{equation}\label{equation: upper bound on KL divergence}
KL(\mathbb{D}_0, \mathbb{D}_i) \leq \sum_{t=1}^T 6 \big(\mathbb{E}_0[N^{(i,1)}_T] + 2^h|L_\mathcal{T}|\alpha T \big) \epsilon^2  \underset{(i)}{\leq} \frac{18T}{M}\epsilon^2 \leq 1~,
\end{equation}
where (i) follows from the definition of $\mathcal{B}$. Finally, using Equations \ref{equation: lower bound in terms of KL Divergence} and \ref{equation: upper bound on KL divergence}, and $|\mathcal{B}|\geq M/2$, we have
\begin{align*}
\sum_{a_{i,1} \in \mathcal{B}} \mathbb{D}_{i}\{a_T \neq a_{i,1}\} &\geq \frac{1}{2} \sum_{a_{i,1} \in \mathcal{B}} \exp(-KL(\mathbb{D}_0, \mathbb{D}_{i}))-1 \nonumber \\
    &\geq \frac{|\mathcal{B}|}{2e} - 1 \nonumber\\
    &\geq \frac{M}{4e} - 1~.
\end{align*}
Therefore as $|\mathcal{B}| \leq M$, by averaging argument there exists an $i \in [M]$ such that 
\begin{align*}
\mathbb{D}_i\{a^*_T \neq a_{i,1}\} \geq \frac{\frac{M}{4e} - 1}{M} ~.   
\end{align*}
\end{proof}

From Lemmas \ref{lemma: lower bound on regret} and \ref{lemma: lower bound on the probability of choosing a sub-optimal arm} for any algorithm \texttt{ALG}, \AUR{if $\epsilon < \frac{1}{4}$} then the  expected simple regret of \texttt{ALG} can be upper bounded as follows
\begin{align}
r_{\texttt{ALG}}(T) \geq \mathbb{D}_i\{a^*_T \neq a_{i,1}\}\frac{1}{2}\epsilon \geq \frac{\frac{M}{4e} - 1}{M} \cdot (\frac{1}{2} \epsilon) \geq \frac{\frac{M}{4e}-1}{2M} \sqrt{\frac{M}{18T}}~.
\end{align}
\AUR{On the contrary, if $\epsilon \geq \frac{1}{4}$ then $M \geq T$, so $\sqrt{M/T} = \Omega(1)$ and regret $r_{\texttt{ALG}}(T) \geq \Omega(1)$}. 
Hence, for any algorithm there exists an $i\in [0,M]$ such that the expected simple regret of the algorithm on $\mathcal{C}_i$ is $\Omega\bigg(\sqrt{\frac{m(\mathcal{C}_i)}{T}} \bigg)$.

\section{Proof of Theorem \ref{theorem: LB-given-q}}\label{secappendix: lower bound for tree given q}

\AUR{We begin by constructing the causal graph $\mathcal{G}$ on $N+1$ nodes $\{X_1, \ldots, X_N, Y\}$, where $N\geq 3$. In $\mathcal{G}$, $X_N$ is the parent of $X_1, \dots, X_{N-1}$ and there is a directed edge form each node to the outcome node $Y$. The strategy remains the same as in the proof of Theorem \ref{theorem: LB-Tree}; Now given $q_1, q_2, \dots, q_N$, compatible with the graph $\mathcal{G}$, we will construct $\mathbb{P}_0, \ldots , \mathbb{P}_{N}$ such that on at least one CBN $\mathcal{C}_i = (\mathcal{G},\mathbb{P}_i)$ the expected simple regret of any algorithm is tight. Also, without loss of generality, assume that $q_1 \leq q_2 \leq \dots \leq q_N$.}

\textbf{Defining} $\mathbb{P}_0$: For all the nodes in the graph $\mathcal{G}$, we define the distribution $\mathbb{P}_0$ as follows:
\begin{align*}
&\mathbb{P}_0(X_N = 1) = q_N \\
&\mathbb{P}_0(X_i = 1 | X_N = 0) = \frac{q_i}{1-q_N} \\
&\mathbb{P}_0(X_i = 1 | X_N = 1) = \frac{1}{2} \\
&\mathbb{P}_0(Y = 1 | . ) = 0.5
\end{align*}

$\mathbb{P}_0(Y = 1 | . )$ denotes the probability of $Y = 1$ conditioned on any value of the parents. Also, note that since $q_1, \dots, q_N$ are compatible with the given graph $\mathcal{G}$, we have, for any $i \neq N$, $q_i = \min_{x_i, x_N} \mathbb{P}_0(X_i = x_i, X_N = x_N) \leq \mathbb{P}_0(X_i = 1, X_N = 1) = q_N/2$.  In addition, $\mathbb{P}_0(X_i = 1 | X_N = 0) = q_i/(1-q_N) \leq 2q_i$. Let $M = m(\mathcal{C}_i)$ for all $i \in [N]$ and $M' = M-1$. 




\textbf{Case a:} $M \geq 12$.

\textbf{Defining} $\mathbb{P}_i$: For $i = N$, define $\mathbb{P}_N(Y = 1 | X_N = 1) = 0.5 + \epsilon$, and for $i \neq N$, $\mathbb{P}_i(Y = 1 | X_i = 1, X_N = 0) = 0.5 + \epsilon$. The remaining conditional distributions are same as those of $\mathbb{P}_0$.

Now, it is easy to see that the optimal action for $\mathbb{P}_i$ is $a_{i,1}$. \VIN{As in proof of Theorem \ref{theorem: LB-Tree}, let $\text{rew}_i(a_{j, x})$ denote the expected reward of action $do(X_j=x)$ under the distribution $\mathbb{P}_i$}.

\begin{lemma}\label{lemma 4.2: LB of regret of sub-optimal arm}
For every $i \in [M']$, $j \in [N]$, $x\in \{0,1\}$, and $(j,x) \neq (i,1)$ the following holds: $\text{rew}_i(a_{i,1}) - \text{rew}_i(a_{j,x}) \geq 0.1\epsilon$.
\end{lemma}
\begin{proof}

For $i=N$, the regret for choosing a sub-optimal arm $a$ is $\text{rew}_N(a_{N,1}) - \text{rew}_N(a) \geq (1-q_{N}) \epsilon \geq 0.5\epsilon$. For $i \neq N$, the regret for choosing a sub-optimal arm $a_{j,x}$, where $j \neq N$ is as follows:
\begin{align}
\text{rew}_i(a_{i,1}) - \text{rew}_i(a_{j,x}) &\geq (1-q_{N}) \epsilon - q_{i}\epsilon \nonumber \\
&\geq \bigg(1-\frac{3q_{N}}{2}\bigg) \epsilon \nonumber \\
&\geq 0.25 \epsilon \nonumber
\end{align}

For $j=N$, the regret is as follows:
\begin{align}
\text{rew}_i(a_{i,1}) - \text{rew}_i(a_{N,0}) &= (1-q_{1}) \epsilon - \mathbb{P}_i(X_i = 1 | X_N = 0)\epsilon \geq (0.5-2q_{i}) \epsilon \nonumber \\
\text{rew}_i(a_{i,1}) - \text{rew}_i(a_{N,1}) &= (1-q_{1}) \epsilon \geq 0.5 \epsilon \nonumber
\end{align}

Hence, if $q_{i} \leq 1/M' \leq \frac{1}{5}$, the regret of pulling an optimal arm is $0.1 \epsilon$.

\end{proof}

Let \texttt{ALG} be an algorithm that outputs arm $a_T$ at the end of $T$ rounds. We choose $\epsilon = \min \{\frac{1}{4}, \sqrt{\frac{M'}{24T}}\}$. For $i \in [N]$, denote $\mathbb{D}_i$ as the measure on all the sampled values of the nodes of $\mathcal{G}$ over $T$ rounds and $\mathbb{E}_i$ as the expectation over $\mathbb{D}_i$. Let $\mathcal{G}_{t}$ be the sampled values of the nodes of $\mathcal{G}$ at time $t$ and let $\mathbf{G}_{t} = \{\mathcal{G}_{1}, \ldots, \mathcal{G}_{t}\}$. Also, for $i \in [0,M']$ let $\mathbb{D}_i(.|\mathbf{G}_{t-1}) = \mathbb{P}_i^t(.)$. Note that \texttt{ALG} determines the arm $a_t$ conditioned on $\mathbf{G}_{t-1}$ (either in a deterministic or randomized way). Also for $j, j' \in [1,N]$, if $a_t = a_{j,x}$ and $j' \neq j$, then $\mathbb{P}_i^t(X_{j'} = x | do(X_j)=x) = \mathbb{P}_i(X_{j'}=x| do(X_j=x))$.

\begin{lemma}\label{lemma 4.2: LB of probability of selecting a sub-optimal arm}
For any algorithm \texttt{ALG}, there exists an $i \in [M']$, such that $\mathbb{D}_i\{a_T \neq a_{i,1}\} \geq \frac{\frac{M'}{4e}-1}{M'}$.
\end{lemma}

\begin{proof}
We use $KL(\mathbb{D}_0, \mathbb{D}_i)$ to denote the KL divergence between $\mathbb{D}_0$ and $\mathbb{D}_i$ for any $i\in [N]$. Let $N^{(i,1)}_T$ be the number of times  \texttt{ALG} plays the arm $a_{i,1}$ at the end of $T$ rounds. Also, let $\mathcal{B} = \{ a_{i,1} \mid  i \leq M' \textit{ and } \mathbb{E}_0[N^{(i,1)}_T] \leq 2T/M' \}$. Observe that $|\mathcal{B}| \geq M'/2$, as otherwise the sum of the expected number of arm pulls of arms not in $\mathcal{B}$ would be greater than $T$. First, using Lemma 2.6 from \cite{10.5555/1522486}, we have, 
\begin{equation*}
\mathbb{D}_0(a_T = a_{i,1}) + \mathbb{D}_i(a_T \neq a_{i,1}) \geq 
\frac{1}{2}\cdot  \exp{(-KL(\mathbb{D}_0, \mathbb{D}_i))} 
\end{equation*}
Rearranging and summing the above equation over arms in $\mathcal{B}$, and observing that $\sum_{a_{i,1} \in \mathcal{B}} \mathbb{D}_0(a_T = a_{i,1}) \leq 1$ we have
\begin{equation}\label{equation: lower bound in terms of KL Divergence 2}
\sum_{a_{i,1} \in \mathcal{B}} \mathbb{D}_{i}\{a_T \neq a_{i,1}\} \geq \frac{1}{2}\cdot \sum_{a_{i,1} \in \mathcal{B}} \exp(-KL(\mathbb{D}_0, \mathbb{D}_{i}))-1 
\end{equation} 
Now we bound $\exp(-KL(\mathbb{D}_0, \mathbb{D}_{i}))$ for every $i$ such that $a_{i,1} \in \mathcal{B}$.
Using the chain rule for product distributions (see \cite{492488} and Chapter 2 in \cite{Slivkins_Book}) the KL divergence of $\mathbb{D}_0$ and $\mathbb{D}_i$ for any $i \in [M]$ can be written as 
\begin{align}
KL(\mathbb{D}_0, \mathbb{D}_i) = \sum_{t=1}^T KL(\mathbb{D}_0(\mathcal{G}_{t} | \mathbf{G}_{t-1}), \mathbb{D}_i(\mathcal{G}_{t} | \mathbf{G}_{t-1}) = \sum_{t=1}^T KL(\mathbb{P}_0^t(\mathcal{G}_t), \mathbb{P}_i^t(\mathcal{G}_t)) \label{kl-divergence-summation}
\end{align}

Now each term in the summation can be written as, for $i \neq N$,
\begin{align}
&KL(\mathbb{P}_0^t, \mathbb{P}_i^t) \nonumber \\
&= \sum_\mathbf{v} \mathbb{P}_0^t(\mathbf{v}) \log{\frac{\mathbb{P}_0^t(\mathbf{v})}{\mathbb{P}_i^t(\mathbf{v})}} \nonumber \\
&= \sum_y \mathbb{P}_0^t(Y = y| X_N = 0, X_i = 1) \mathbb{P}_0^t(X_N = 0, X_i = 1) \log{\frac{\mathbb{P}_0^t(Y = y | X_N = 0, X_i = 1)}{\mathbb{P}_i^t(Y = y | X_N = 0, X_i = 1)}} \nonumber \\
&= 0.5\mathbb{P}_0^t(X_N = 0, X_i = 1) \bigg[ \log{\frac{0.5}{0.5 + \epsilon}} + \log{\frac{0.5}{0.5 - \epsilon}} \bigg] \nonumber \\
&\leq 6\mathbb{P}_0^t(X_N = 0, X_i = 1)\epsilon^2 \label{kl-i-neq-1}
\end{align}

For $i = N$,
\begin{align}
KL(\mathbb{P}_0^t, \mathbb{P}_i^t) &= \sum_\mathbf{v} \mathbb{P}_0^t(\mathbf{v}) \log{\frac{\mathbb{P}_0^t(\mathbf{v})}{\mathbb{P}_i^t(\mathbf{v})}} \nonumber \\
&= \sum_y \mathbb{P}_0^t(Y = y | X_N = 1) \mathbb{P}_0^t(X_N = 1) \log{\frac{\mathbb{P}_0^t(Y = y | X_1 = 1)}{\mathbb{P}_i^t(Y = y | X_N = 1)}} \nonumber \\
&= 0.5\mathbb{P}_0^t(X_N = 1) \bigg[ \log{\frac{0.5}{0.5 + \epsilon}} + \log{\frac{0.5}{0.5 - \epsilon}} \bigg] \nonumber \\
&\leq 6\mathbb{P}_0^t(X_N = 1)\epsilon^2 \label{kl-i-eq-1}
\end{align}

Using Equation \ref{kl-i-neq-1} and \ref{kl-i-eq-1} in equation \ref{kl-divergence-summation}, we get when $q_i \leq \frac{1}{M'}$
\begin{align*}
KL(\mathbb{D}_0, \mathbb{D}_i) &\leq 6 \bigg[ \mathbb{E}_0[N_T^{(i,1)}] + \frac{2}{M'}T \bigg] \epsilon^2 \\
&\leq \frac{24T}{M'} \epsilon^2 \\
&\leq 1 \\
\end{align*}

Now putting the value of $KL(\mathbb{D}_0, \mathbb{D}_i)$ in Equation \ref{equation: lower bound in terms of KL Divergence 2} we get the following,
\begin{align*}
\sum_{a_{i,1} \in \mathcal{B}} \mathbb{D}_{i}\{a_T \neq a_{i,1}\} &\geq \frac{1}{2} \sum_{a_{i,1} \in \mathcal{B}} \exp(-KL(\mathbb{D}_0, \mathbb{D}_{i}))-1 \nonumber \\
    &\geq \frac{|\mathcal{B}|}{2e} - 1 \nonumber\\
    &\geq \frac{M'}{4e} - 1~.
\end{align*}
Therefore as $|\mathcal{B}| \leq M'$, by averaging argument there exists an $i \in [M']$ such that 
\begin{align*}
\mathbb{D}_i\{a^*_T \neq a_{i,1}\} \geq \frac{\frac{M'}{4e} - 1}{M'} ~.   
\end{align*}

From Lemmas \ref{lemma 4.2: LB of regret of sub-optimal arm} and \ref{lemma 4.2: LB of probability of selecting a sub-optimal arm} for any algorithm \texttt{ALG}, \AUR{if $\epsilon < \frac{1}{4}$} then the  expected simple regret of \texttt{ALG} can be upper bounded as follows
\begin{align} \label{M-greater-than=5}
r_{\texttt{ALG}}(T) \geq \mathbb{D}_i\{a^*_T \neq a_{i,1}\}\cdot (0.1\epsilon) \geq \frac{\frac{M'}{4e} - 1}{M'} \cdot (0.1 \epsilon) \geq \frac{\frac{M'}{4e}-1}{10M'} \sqrt{\frac{M'}{24T}}~.
\end{align}

\AUR{Otherwise, if $\epsilon \geq \frac{1}{4}$, $M' \geq T$, so $\sqrt{M'/T} = \Omega(1)$ and regret $r_{\texttt{ALG}}(T) \geq \Omega(1)$}.

Hence, it is proved that regret is lower bounded by $\Omega\big(\sqrt{\frac{M}{T}}\big)$.

\end{proof}

\textbf{Case b:} $M < 12$. Define $N$ distributions $\mathbb{P}_1, \dots, \mathbb{P}_N$ as follows. We choose $\epsilon = \sqrt{\frac{1}{45T}}$. The rest of conditional distributions remain same as $\mathbb{P}_0$. For all $i \in [N]$, 
\begin{align*}
&\mathbb{P}_i(Y = 1 | X_i = 1) = 0.5 + \epsilon
\end{align*}

Now, the optimal arm for action $\mathbb{P}_i$ is $a_{i,1}$, and the regret of pulling a sub-optimal arm in place of the optimal arm $a_{i,1}$ is $(1-q_i) \epsilon \geq 0.5\cdot\epsilon$. Each term in the summation of Equation \ref{kl-divergence-summation} can be written as
\begin{align*}
KL(\mathbb{P}_0^t, \mathbb{P}_i^t) &= \sum_\mathbf{v} \mathbb{P}_0(\mathbf{v}) \log{\frac{\mathbb{P}_0^t(\mathbf{v})}{\mathbb{P}_i^t(\mathbf{v})}} \\
&= \sum_y \mathbb{P}_0^t(Y = y | X_i = 1) \mathbb{P}_0^t(X_i = 1) \log{\frac{\mathbb{P}_0^t(Y = y| X_i = 1)}{\mathbb{P}_i^t(Y = y | X_i = 1)}} \\
&= 0.5\mathbb{P}_0^t(X_i = 1) \bigg[ \log{\frac{0.5}{0.5 + \epsilon}} + \log{\frac{0.5}{0.5 - \epsilon}} \bigg] \\
&\leq 6\mathbb{P}_0^t(X_i = 1)\epsilon^2
\end{align*}

Since $\mathbb{P}_0(X_i = 1 | .) \leq 0.5$. 
\begin{align}
KL(\mathbb{D}_0, \mathbb{D}_i) &\leq 6 \bigg[ \mathbb{E}_0[N_T^{(i,1)}] + \frac{T}{2} \bigg] \epsilon^2
\end{align}

Note that $\mathbb{E}_0[N_T^{(i,1)}] \leq T$ 
\begin{align}
KL(\mathbb{D}_0, \mathbb{D}_i) &\leq 9T \epsilon^2 \leq 0.2
\end{align}

Now putting the value of $KL(\mathbb{D}_0, \mathbb{D}_i)$ in Equation \ref{equation: lower bound in terms of KL Divergence 2} we get the following,
\begin{align*}
\sum_{i \in [N]} \mathbb{D}_{i}\{a_T \neq a_{i,1}\} &\geq \frac{1}{2} \sum_{i \in [N]}  \exp(-KL(\mathbb{D}_0, \mathbb{D}_{i}))-1 \nonumber \\
    &\geq \frac{N}{2e^{0.2}} - 1~.
\end{align*}

Hence  any algorithm \texttt{ALG} there exists an $i$ such that the regret incurred by it is
\begin{align}
r_{\texttt{ALG}}(T) &\geq 0.5 \mathbb{D}_i(a_T \neq a_{i,1}) \epsilon \geq \frac{\frac{N}{2e^{0.2}}-1}{N} \sqrt{\frac{1}{45T}}  
\label{M-less-than-6}
\end{align}

Finally, from Equations \ref{M-greater-than=5} and \ref{M-less-than-6} it follows that the expected simple regret of any algorithm is $\Omega\big(\sqrt{\frac{M}{T}}\big)$, where $M$ depends on $\mathbf{q}$ and $k_i$ for $i \in [N]$.
\section{Proof of Theorem \ref{theorem: UB-CRM}}\label{secappendix: proof of CRM}
Throughout the proof we use $a^*$ to denote the optimal arm. First, we prove a few lemmas, and then use it to bound the expected cumulative regret of \CRM. The following lemma shows that the expectation of $\widehat{\mu}_{i,x}$ as defined in Equation \ref{equation: emprical estimate for arm i,x modified} is equal to $\mu_{i,x}$ for every $i,x$.
\begin{lemma}\label{lemma: unbiased muix}
$\widehat{\mu}_{i,x}(t)$ is an unbiased estimator of $\mu_{i,x}$, that is $\mathbb{E}[\widehat{\mu}_{i,x}(t)] = \mu_{i,x}$. Moreover $\mathbb{P}(|\widehat{\mu}_{i,x}(t) - \mu_{i,x}| \geq \epsilon) \leq 2\exp(-2(N_t^{i,x} + C_t^{i,x})\epsilon^2)$~.
\end{lemma}
\begin{proof}
We begin by restating the the definition of $\widehat{\mu}_{i,x}$ from  Equation \ref{equation: emprical estimate for arm i,x modified}.
\begin{equation*}
    \widehat{\mu}_{i,x}(t) = \frac{\sum_{j \in S_t^{i,x}}\mathds{1}\{Y_j=1\} + \sum_{c \in [C_t^{i,x}]} Y_c^{i,x}}{N^{i,x}_t + C^{i,x}_t}
\end{equation*}
We note that in Equation \ref{equation: emprical estimate for arm i,x modified}, $Y_c^{i,x}$ is a random variable such that $\mathbb{E}[Y_c^{i,x}] = \mu_{i,x}$. Note that this holds because we partition the time steps where arm $a_0$ was pulled into odd and even instances $S^0_{o,t}$ and $S^{0}_{e,t}$.Taking expectation on both sides of the above equation we have
\begin{align*}
&\mathbb{E}[\widehat{\mu}_{i,x}(t)] \nonumber \\
&= \mathbb{E}\bigg[ \frac{\sum_{j \in S_t^{i,x}}\mathds{1}\{Y_j=1\} + \sum_{c \in [C_t^{i,x}]} Y_c^{i,x}}{N^{i,x}_t + C^{i,x}_t} \bigg] \nonumber \\
&= \sum_{a=1}^{\infty} \sum_{b=0}^{\infty} \mathbb{E}\bigg[ \frac{\sum_{j \in S_t^{i,x}}\mathds{1}\{Y_j=1\} + \sum_{c \in [C_t^{i,x}]} Y_c^{i,x}}{N^{i,x}_t + C^{i,x}_t} \Bigl\vert N_t^{i,x} = a, C_t^{i,x} = b \bigg] \mathbb{P}(N_t^{i,x} = a, C_t^{i,x} = b) \nonumber \\
&= \sum_{a=1}^{\infty} \sum_{b=0}^{\infty} \bigg(\frac{a\mu_{i,x} + b\mu_{i,x}}{a + b}\bigg) \mathbb{P}(N_t^{i,x} = a, C_t^{i,x} = b) \nonumber \\
&= \mu_{i,x} \sum_{a=1}^{\infty} \sum_{b=0}^{\infty} \mathbb{P}(N_t^{i,x} = a, C_t^{i,x} = b) \nonumber \\
&= \mu_{i,x}
\end{align*}

Next we prove the concentration inequality part of the lemma, which is similar to Chernoff-Hoeffding inequality (Lemma \ref{lemma: chernoff-hoeffding inequality}) for our estimator. 

\begin{align}
&\mathbb{P}\bigg(\frac{\sum_{j\in S_t^{i,x}} \mathds{1}\{Y_j = 1\} + \sum_{c\in [C^{i,x}_t]}Y^{i,x}_c}{N_t^{i,x} + C_t^{i,x}} \geq \mu_{i,x} + \epsilon \bigg) \nonumber \\
&= \mathbb{P}\bigg(\sum_{j\in S_t^{i,x}} \mathds{1}\{Y_j = 1\} + \sum_{c\in [C^{i,x}_t]}Y^{i,x}_c \geq (N_t^{i,x} + C_t^{i,x})\mu_{i,x} + (N_t^{i,x} + C_t^{i,x})\epsilon \bigg) \nonumber \\
&\stackrel{(i)}{\leq} \min_{\lambda \geq 0} E\bigg[\exp{\Big(\lambda\big(\sum_{j\in S_t^{i,x}} (\mathds{1}\{Y_j = 1\} - \mu_{i,x}) + \sum_{c \in [C_t^{i,x}]}(Y^{i,x}_c - \mu_{i,x}) \big) \Big)} \bigg]e^{-\lambda(N_t^{i,x} + C_t^{i,x})\epsilon} \nonumber \\
&= \min_{\lambda \geq 0} E\bigg[ \prod_{j \in S_t^{i,x}} \exp \big(\lambda (\mathds{1}\{Y_j = 1\} - \mu_{i,x}) \big) \prod_{c \in [C_t^{i,x}]} \exp \big(\lambda (Y_c^{i,x} - \mu_{i,x}) \big) \bigg]e^{-\lambda(N_t^{i,x} + C_t^{i,x})\epsilon} \nonumber \\
&\stackrel{(ii)}{=} \min_{\lambda \geq 0} \prod_{j \in S_t^{i,x}} E\bigg[\exp \big(\lambda (\mathds{1}\{Y_j = 1\} - \mu_{i,x}) \big) \bigg] \prod_{c \in [C_t^{i,x}]} E\bigg[ \exp \big(\lambda (Y_c^{i,x} - \mu_{i,x}) \big) \bigg]e^{-\lambda(N_t^{i,x} + C_t^{i,x})\epsilon} \nonumber \\
&\stackrel{(iii)}{\leq} \min_{\lambda \geq 0} \exp \bigg(\frac{N_T^{i,x}\lambda^2}{8} + \frac{C_t^{i,x}\lambda^2}{8} - \lambda(N_T^{i,x} + C^{i,x}_t) \epsilon\bigg) \nonumber \\
&\leq \exp{(-2(N_t^{i,x} + C^{i,x}_t)\epsilon^2)}
\end{align}
In the above equations, the inequality in $(i)$ follows from Lemma \ref{lemma: Chernoff bound}, the equality in $(ii)$ follows from the fact that each term in the product are independent, and $(iii)$ follows from Lemma \ref{lemma: Hoeffding's lemma}. We use that each $|S^{i,x}_{\mathbf{z},t}| = C^{i,x}_{t}$ in $(i)$ and it is for this step we truncate $S^{i,x}_{\mathbf{z},t}$ to $C^{i,x}_{t}$ elements. Following the same steps as above we get the following two sided bound
\begin{align}
    \mathbb{P}(|\widehat{\mu}_{i,x}(t) - \mu_{i,x}| \geq \epsilon) \leq 2\exp{(-2(N_t^{i,x} + C_t^{i,x})\epsilon^2)} \ .
\end{align}
\end{proof}

Next we show that the estimates of $\mu_a$ at the end of $T$ rounds is good with high probability.

\begin{lemma} \label{lemma: concentration bounds on mu in cumulative regret}
Let $p = \min_{i,x,\mathbf{z}} \mathbb{P}(X_i = x, \mathbf{Pa}(X_i) = \mathbf{z})$.  Then for any $T \in \mathbb{N}$, at the end of $T$ rounds the following hold:  
\begin{enumerate}
    \item $\mathbb{P} \Big\{|\widehat{\mu}_0(T) - \mu_0| \geq \frac{\Delta_0}{4} \Big\} \leq  2T^{-\frac{\Delta_0^2}{8}}$~.
    \item Let $\widehat{p}^{\ i,x}_{\mathbf{z}, T} = \frac{\sum_{t \in S_{o,T}^0} \mathds{1}\{a_t = a_0, X_i=x, \mathbf{Pa}(X_i)=\mathbf{z}\}}{|S_{o,T}^0|}$, and $\widehat{p}^{\ i,x}_T = \min_\mathbf{z} \widehat{p}^{\ i,x}_{\mathbf{z},T}$. Then
    $\mathbb{P}\{\widehat{p}^{i,x}_T \geq \frac{p}{2} \} \geq  1-Z_iT^{-\frac{p^2}{4}}$, where $Z_i$ is the size of the domain from which $\mathbf{Pa}(X_i)$ takes values.
    \item $\mathbb{P}\Big\{|\widehat{\mu}_{i,x}(T) - \mu_{i,x}| \geq \frac{\Delta_0}{4} \Big\} \leq 2T^{-\frac{p\Delta_0^2}{32}} + Z_iT^{-\frac{p^2}{4}}$~.
\end{enumerate}
\end{lemma}
\begin{proof}
a) Since $\beta\geq 1$, at the end of $T$ rounds arm $a_0$ is pulled by Algorithm \ref{CR-algorithm} at least $(\ln{T})$ times. Hence, $N^0_T \geq (\ln{T})$, and by \ref{lemma: chernoff-hoeffding inequality},
\begin{equation*}\label{equation: empirical estimate of mu_0 is close}
\mathbb{P}\Big\{|\widehat{\mu}_0(T) - \mu_0| \geq \frac{\Delta_0}{4} \Big\} \leq 2e^{-\frac{\Delta_0^2}{8}\ln T} = 2T^{-\frac{\Delta_0^2}{8}}
\end{equation*}

b) \AUR{In this part we show, using union bound, that the estimation of $\widehat{p}^{\ i,x}_{T}$ being less that $p/2$ have low probability.}
Since, $|S_{o,T}^0| \geq N_T^0/2$, by Lemma \ref{lemma: chernoff-hoeffding inequality}, we have,
\begin{equation*}
\mathbb{P}\bigg(\widehat{p}^{i,x}_{\mathbf{z}, T} > p^{i,x}_{\mathbf{z}} - \frac{p}{2} \geq \frac{p}{2}\bigg) \geq 1 - e^{-2 \frac{p^2}{4} \frac{\ln T}{2}} = 1 - T^{-\frac{p^2}{4}} 
\end{equation*}
Now using this we get,
\begin{align}
\label{p-ix-greater-p-2}
\mathbb{P}\bigg(\widehat{p}^{\ i,x}_{T} \leq \frac{p}{2}\bigg) =  \mathbb{P}\bigg(\min_\mathbf{z} \widehat{p}^{\ i,x}_{\mathbf{z}, T} \leq \frac{p}{2}\bigg) \leq \sum_\mathbf{z} \mathbb{P}\bigg(\widehat{p}^{\ i,x}_{\mathbf{z}, T} \leq \frac{p}{2}\bigg) \leq Z_iT^{-\frac{p^2}{4}}
\end{align}

c) Let the conditional probability distribution $\mathbb{P}(.| \widehat{p}^{\ i,x}_T > \frac{p}{2})$ be denoted by $\mathbb{P}_p$. Since $\beta \geq1$, $N_T^0 \geq \ln{T}$. Further if $\widehat{p}^{\ i,x}_T > \frac{p}{2}$ then $C^{i,x}_T > \frac{p}{2} \frac{N^0_T}{2} \geq \frac{p}{4} \ln T$ (from the definition of $C^{i,x}_T$). Hence, from Lemma \ref{lemma: unbiased muix} we have
\begin{equation}
    \label{bounding-mu-ix-delta}
    \mathbb{P}_p\bigg(|\widehat{\mu}_{i,x}(T) - \mu_{i,x}| \geq \frac{\Delta_0}{4}\bigg) \leq 2\exp\bigg(-\frac{\Delta_0^2}{32}p\ln T\bigg) = 2T^{-\frac{p\Delta_0^2}{32}}
\end{equation}
Finally by the law of total probability and using Equations \ref{p-ix-greater-p-2} and  \ref{bounding-mu-ix-delta}
\begin{align*}
\mathbb{P}\bigg(|\widehat{\mu}_{i,x}(T) - \mu_{i,x}| \geq \frac{\Delta_0}{4}\bigg) &\leq \mathbb{P}_p\bigg(|\widehat{\mu}_{i,x}(T) - \mu_{i,x}| \geq \frac{\Delta_0}{4}\bigg) + \mathbb{P}\bigg(\widehat{p}^{\ i,x}_T \leq \frac{p}{2}\bigg) \nonumber \\
&\leq 2T^{-\frac{p\Delta_0^2}{32}} + Z_iT^{-\frac{p^2}{4}}
\end{align*}
\end{proof}

Next we show that $\beta$ as set in \CRM\ is bounded in expectation. Lemma \ref{bounding E[beta^2]} and its proof is similar to Lemma 8.6 in \cite{NairPS21}.
\begin{lemma} \label{bounding E[beta^2]}
Let $L = \arg \min_{t \in \mathds{N}} \bigg\{\frac{t^\frac{p^2\Delta_0^2}{32}}{\ln{t}} \geq 3N(Z+3)\bigg\}$, where $Z = \max_{i} Z_i$, and suppose \CRM\ pulls arms for T rounds, where $T \geq \max(L, e^{\frac{50}{\Delta_0^2}})$, and let $a^* \neq a_0$. Then at the end of T rounds, $\frac{8}{9\Delta_0^2} \leq \mathbb{E}[\beta^2] \leq \frac{50}{\Delta_0^2}$.
\end{lemma}
\begin{proof}
Before proceeding to the proof of the lemma we make the following two observations.
\begin{observation}
1. If $a^{*} \neq a_0$ then $\Delta_0 = \mu_{a^*} - \mu_0$ \\
2. Let $\widehat{\mu}^* = \max_{i,x}(\widehat{\mu}_{i,x}(T))$. If $|\widehat{\mu}_0(T) - \mu_0| \leq \frac{\Delta_0}{4}$ and $|\widehat{\mu}_{i,x}(T) - \mu_{i,x}| \leq  \frac{\Delta_0}{4}$ for all $(i,x)$ then $\frac{\Delta_0}{2} \leq \widehat{\mu}_{a^*} - \widehat{\mu}_0(T) \leq \frac{3\Delta_0}{2}$, and $\frac{32}{9\Delta_0^2} \leq \beta^2 \leq \frac{32}{\Delta_0^2}$. Notice that since $T\geq e^{\frac{50}{\Delta_0^2}}$, $\frac{32}{\Delta_0^2} \leq \ln T$.
\end{observation}

Let $U_0$ be the event that $|\widehat{\mu}_0(T) - \mu_0| \leq \frac{\Delta_0}{4}$, and for any $i,x$ let $U_{i,x}$ be the event $|\widehat{\mu}_{i,x}(T) - \mu_{i,x}| \leq \frac{\Delta_0}{4}$. Also let $U = (\cap_{i,x} U_{i,x}) \cap U_0$. If $\overline{U}_0$, $\overline{U}_{i,x}$, and $\overline{U}$ denote the compliment of the events $U_0, U_{i,x}$, and $U$ respectively, then

$$
\mathbb{P}\{\overline{U}_0\} \leq 2T^{-\frac{\Delta_0^2}{8}}~, \text{~and}$$
$$\text{for a fixed } (i,x)~~~\mathbb{P}\{ \overline{U}_{i,x} \} \leq 2T^{-\frac{p\Delta_0^2}{32}} + Z_iT^{-\frac{p^2}{4}}~.
$$

Hence applying union bound, 
\begin{align*}
\mathbb{P}\{\overline{U}\} &\leq 2N\left(\frac{2}{T^{\frac{p\Delta_0^2}{32}}} +  \frac{Z}{T^{\frac{p^2}{4}}}\right) + \frac{2}{T^{\frac{\Delta_0^2}{8}}} \nonumber \\
&\leq 2N\left(\frac{2}{T^{\frac{p^2\Delta_0^2}{32}}} +  \frac{Z}{T^{\frac{p^2\Delta_0^2}{32}}}\right) + \frac{2N}{T^{\frac{p^2\Delta_0^2}{32}}} ~~~~~~~~~~~~\text{as }~p\leq 1, \Delta_0\leq 1 \nonumber \\
&\leq \frac{2N(Z+3)}{T^{\frac{p^2\Delta_0^2}{32}}} =\delta
\end{align*}

We will use the above arguments to first show that $\mathbb{E}[\beta^2] \geq \frac{8}{\Delta_0^2}$. From part 2 of Observation we have that the event $U$ implies $\beta^{2} \geq \frac{32}{9\Delta_0^2}$. Since $\mathbb{P}\{U\} \geq 1-\delta$,
$$\mathbb{E}[\beta^{2}] \geq \frac{32}{9\Delta_0^2} (1-\delta) = \frac{32}{9\Delta_0^2} - \frac{32\delta}{9\Delta_0^2} $$
Since $T$ satisfies $\frac{T^{\frac{p^2\Delta_0^2}{32}}}{\ln T} \geq 3N(Z + 3)$, this implies $\frac{32\delta}{9\Delta_0^2} \leq \frac{24}{9\Delta_0^2}$, and hence $\mathbb{E}[\beta^2] \geq \frac{8}{9\Delta_0^2}$. Similarly, from part 2 of Observation we have that the event $U$ implies $\beta^{2} \leq \frac{32}{\Delta_0^2}$. If $U$ does not hold then $\beta^2 \leq \ln T$. Hence using the fact that $T$ satisfies $\frac{T^{\frac{p^2\Delta_0^2}{32}}}{\ln T} \geq 3N(Z+3)$, and hence $\delta \ln T \leq \frac{18}{\Delta_0^2}$, we get,
$$\mathbb{E}[\beta^{2}] \leq \frac{32}{\Delta_0^2} (1-\delta) + \delta\ln T \leq \frac{32}{\Delta_0^2} + \delta \ln T \leq \frac{50}{\Delta_0^2}~.$$

\end{proof}

\begin{lemma} \label{lemma: bound E[N_T^ix]}
Suppose $a^* \neq a_{i,x}$. Then at the end of $T$ rounds the following holds:
$$\mathbb{E}[N^{i,x}_{T}] \leq \max\left(0, \frac{8\ln T}{\Delta_{i,x}^2} + 1 - \frac{1}{4} \cdot p_{i,x} \cdot \eta_T^{i,x} \cdot \mathbb{E}[N_T^0]\right) + \frac{\pi^2}{3}~.$$
Further if $a^* \neq a_0$ then
$$\mathbb{E}[N^{0}_T] \leq \Big( \mathbb{E}[\beta^2]\ln T + ~\frac{8\ln T}{\Delta_{0}^2} + 1\Big) + \frac{\pi^2}{3} ~.$$
\end{lemma}
\begin{proof}
Let $E_T^{i,x} = N_T^{i,x} + C_T^{i,x}$. Then, 
\begin{equation}\label{equation: value of N^i,x_T}
    N^{i,x}_{T} = \sum_{t\in T} \mathds{1}\{a_t = a_{i,x}\} ~.
\end{equation}
\begin{align}
N_T^{i,x} \leq \max(0, \ell - C_T^{i,x}) + \sum_{t \in T} \mathds{1}\{a_t = a_{i,x}, E_t^{i,x}\geq \ell\} \label{N-T-i-x-l}
\end{align}

Here, we make an observation regarding the expected value of $C_T^{i,x}$.
\begin{observation}
$\mathbb{E}[C_T^{i,x}] = \mathbb{E}[\min_\mathbf{z} \widehat{p}^{i,x}_{\mathbf{z},T} \lceil N_T^0/2 \rceil] \geq \frac{1}{4} \cdot p_{i,x} \cdot \mathbb{E}[N_T^0] \cdot (1 - Z_i T^{-\frac{p_{i,x}^2}{2}}) = \frac{1}{4} \cdot p_{i,x} \cdot \eta_T^{i,x} \cdot \mathbb{E}[N_T^0]$
\end{observation}
\begin{proof}
Note that the expectation of $\min_{\mathbf{z}}\widehat{p}^{i,x}_{\mathbf{z},T}$ is over the distribution of the CBN and that of $N_T^0$ over the distribution in the observation across all $T$ rounds. Recall $p_{i,x} = \min_{\mathbf{z}} p_{\mathbf{z}}^{i,x}$. By Lemma \ref{lemma: chernoff-hoeffding inequality}, we have,
\begin{equation*}
\mathbb{P}\bigg(\widehat{p}^{\ i,x}_{\mathbf{z}, T} > p^{i,x}_{\mathbf{z}} - \frac{p_{i,x}}{2} \geq \frac{p_{i,x}}{2}\bigg) \geq 1 - e^{-2 \frac{p_{i,x}^2}{4} \frac{\ln T}{2}} = 1 - T^{-\frac{p_{i,x}^2}{4}} 
\end{equation*}
Now using this we get,
\begin{align*}
\mathbb{P}\bigg(\widehat{p}^{\ i,x}_{T} \leq \frac{p_{i,x}}{2}\bigg) \leq  \mathbb{P}\bigg(\min_\mathbf{z} \widehat{p}^{\ i,x}_{\mathbf{z}, T} \leq \frac{p_{i,x}}{2}\bigg) \leq \sum_\mathbf{z} \mathbb{P}\bigg(\widehat{p}^{\ i,x}_{\mathbf{z}, T} \leq \frac{p_{i,x}}{2}\bigg) \leq Z_iT^{-\frac{p_{i,x}^2}{4}}
\end{align*}
We can now bound the expectation of $C_T^{i,x}$ as follows:
\begin{align}
\mathbb{E}[\min_\mathbf{z} \widehat{p}^{i,x}_{\mathbf{z},T} \lceil N_T^0/2 \rceil] 
&\geq \frac{1}{2} \mathbb{E}[\min_{\mathbf{z}} \widehat{p}^{i,x}_{\mathbf{z},T} N_T^0] \nonumber \\
&= \frac{1}{2} \sum_{a=1}^{\infty} a\cdot\mathbb{E}[\min_{\mathbf{z}} \widehat{p}^{i,x}_{\mathbf{z},T} \mid N_T^0 = a] \mathbb{P}(N_T^0 = a) \nonumber \\
&\geq \frac{1}{2} \sum_{a=1}^{\infty} a \cdot\frac{p_{i,x}}{2} \cdot \mathbb{P}\bigg(\min_\mathbf{z} \widehat{p}^{i,x}_{\mathbf{z},T} > \frac{p_{i,x}}{2} \mid N_T^0 = a\bigg) \mathbb{P}(N_T^0 = a) \nonumber \\
&\geq \frac{1}{2} \sum_{a=1}^{\infty} a \cdot\frac{p_{i,x}}{2} \cdot \mathbb{P}\bigg(\min_\mathbf{z} \widehat{p}^{i,x}_{\mathbf{z},T} > \frac{p_{i,x}}{2} \mid N_T^0 = a\bigg) \mathbb{P}(N_T^0 = a) \nonumber \\
&\geq \frac{p_{i,x}}{4} \mathbb{E}[N_T^0] \cdot \max(0, 1 - Z_i T^{-\frac{p_{i,x}^2}{4}}) \nonumber \\
&= \frac{1}{4} \cdot p_{i,x} \cdot \eta_T^{i,x} \cdot \mathbb{E}[N_T^0]
\end{align}
\end{proof}

Taking expectation of Equation \ref{N-T-i-x-l}, we get
\begin{align} \label{equation: bounding the expected number of pulls of sub-optimal arm}
\mathbb{E}[N_T^{i,x}] \leq \max(0, \ell - \frac{p_{i,x}}{4} \cdot \eta_T^{i,x} \cdot \mathbb{E}[N_T^0] ) + \sum_{t \in [\ell+1, T]} \mathbb{P}\{a_t = a_{i,x}, \mathbb{E}_t^{i,x}\geq \ell\}
\end{align}

Now we bound $\sum_{t\in [l+1,T]} \mathbb{P}\{a(t) = a_{i,x}, E^{i,x}_{t} \geq \ell\}$, and assuming $a^*\neq a_0$. The proof for $a^*=a_0$ is similar. We use $E^{a^*}_T$ to denote the effective number of pulls of $a^*$ at the end of $T$ rounds. Also, for better clarity, we use $\widehat{\mu}_{i,x}(E^{i,x}_T,T)$ (instead of $\widehat{\mu}_{i,x}(T)$) and $\widehat{\mu}_{0}(N^{0}_T,T)$ (instead of $\widehat{\mu}_{0}(T)$) to denote the empirical estimates of $\mu_{i,x}$ and $\mu_0$  computed by Algorithm \ref{CR-algorithm} at the end of $T$ rounds.

\begin{align*}
&\sum_{t\in [\ell+1,T]} \mathbb{P}\Bigg\{a_t = a_{i,x}, E^{i,x}_{t} \geq \ell\Bigg\} \\
&= \sum_{t\in [\ell,T-1]} \mathbb{P}\Bigg\{\widehat{\mu}_{a^*}(E^{a^*}_t,t) + \sqrt{\frac{2\ln t}{E^{a^*}_{t}}} \leq \widehat{\mu}_{i,x}(E^{i,x}_t, t) + \sqrt{\frac{2\ln (t)}{E^{i,x}_{t}}},~~ E^{i,x}_{t} \geq \ell \Bigg\} \\
&\leq  \sum_{t\in [0,T-1]} \mathbb{P}\Bigg\{ \text{min}_{s\in [0,t]}\widehat{\mu}_{a^*}(s,t) + \sqrt{\frac{2\ln t}{s}} \leq \text{max}_{s_j\in [\ell-1,t]}\widehat{\mu}_{i,x}(s_j,t) + \sqrt{\frac{2\ln t}{s_j}} \Bigg\} \\
&\leq \sum_{t\in [T]} \sum_{s\in [0,t-1]} \sum_{s_j \in [\ell-1,t]} \mathbb{P}\Bigg\{\widehat{\mu}_{a^*}(s,t) + \sqrt{\frac{2\ln t}{s}} \leq \widehat{\mu}_{i,x}(s_j, t) + \sqrt{\frac{2\ln t}{s_j}}\Bigg\}
\end{align*}

If $\widehat{\mu}_{a^*}(s,t) + \sqrt{\frac{2\ln t}{s}} \leq \widehat{\mu}_{i,x}(s_j,t) + \sqrt{\frac{2\ln t}{s_j}}$ is true then at least one of the following events is true
\begin{subequations}
\begin{align}
   \widehat{\mu}_{a^*}(s,t) &\leq \mu_{a^*} - \sqrt{\frac{2\ln t}{s}} ~, \label{equation: ucb event a}\\
   \widehat{\mu}_{i,x}(s_j,t) &\geq \mu_{i,x} + \sqrt{\frac{2\ln t}{s_j}} ~, \label{equation: ucb event b}\\
   \mu_{a^*} &\leq \mu_{i,x} + 2\sqrt{\frac{2\ln t}{s_j}}~. \label{equation: ucb event c}
\end{align}
\end{subequations}
The probability of the events in Equations \ref{equation: ucb event a} and \ref{equation: ucb event b} can be bounded using Chernoff-Hoeffding inequality
$$\mathbb{P}\Bigg\{\widehat{\mu}_{a^*}(s,t) \leq \mu_{a^*} - \sqrt{\frac{2\ln t}{s}}\Bigg\} \leq t^{-4} ~,$$ 
$$\mathbb{P}\Bigg\{\widehat{\mu}_{i,x}(s_j,t) \geq \mu_{i,x} + \sqrt{\frac{2\ln t}{s_j}}\Bigg\} \leq t^{-4} ~.$$

Also if $\ell \geq \lceil \frac{8\ln T}{\Delta_{i,x}^2} \rceil$  then the event in Equation \ref{equation: ucb event c} is false, i.e. $\mu_{a^*} >  \mu_{i,x} + 2\sqrt{\frac{2\ln t}{s_j}}$. Thus for $\ell = \frac{8\ln T}{\Delta_{i,x}^2} + 1 \geq \lceil\frac{8\ln T}{\Delta_{i,x}^2}\rceil$, which implies
\begin{equation}\label{equation: probability of a suboptimal pull}
\sum_{t\in [\ell+1,T]} \mathbb{P}\{a(t) = a_{i,x}, E^{i,x}_{t} \geq \ell\} \leq \sum_{t\in [T]}\sum_{s\in [0,t-1]}\sum_{s_j\in [\ell-1, t]} 2t^{-4} \leq \frac{\pi^2}{3}
\end{equation}

If $a^* = a_0$ then using the exact arguments as above we can show that Equation \ref{equation: probability of a suboptimal pull} still holds. Hence, using Equations \ref{equation: bounding the expected number of pulls of sub-optimal arm} and \ref{equation: probability of a suboptimal pull} we have if $a^* \neq a_{i,x}$ then
$$\mathbb{E}[N^{i,x}_{T}] \leq \text{max}\left(0, \frac{8\ln T}{\Delta_{i,x}^2} + 1 - \frac{p_{i,x}}{4} \cdot \eta_T^{i,x} \cdot  \mathbb{E}[N_T^0]\right) + \frac{\pi^2}{3} ~.$$
The arguments used to bound $\mathbb{E}[N^0_T]$, when $a^*\neq a_0$ is similar. In this case the equation corresponding to Equation \ref{equation: bounding the expected number of pulls of sub-optimal arm} is 
\begin{equation}\label{equation: bounding the expected number of pulls of sub-optimal arm a_0}
    \mathbb{E}[N^{0}_{T}] \leq \mathbb{E}[\beta^2]\ln T + \ell + \sum_{t\in [\ell+1, T]} \mathbb{P}\{a(t) = a_0, N^{0}_{t} \geq \ell\} ~.
\end{equation}
Also the same arguments as above can be used to show that for $\ell =  \frac{8\ln T}{\Delta_{0}^2} + 1$,
\begin{equation}\label{equation: probability of a suboptimal pull a_0}
\sum_{t\in T} \mathbb{P}\{a(t) = a_{0}, N^{0}_{t} \geq \ell\}  \leq \frac{\pi^2}{3}~.
\end{equation}
Finally using Equations \ref{equation: bounding the expected number of pulls of sub-optimal arm a_0} and \ref{equation: probability of a suboptimal pull a_0}, we have
$$\mathbb{E}[N^{0}_{T}] \leq \left(\mathbb{E}[\beta^{2}]\ln T + \frac{8\ln T}{\Delta_{0}^2} + 1 \right)  + \frac{\pi^2}{3}~.$$

\begin{lemma} \label{lemma: bound E[N_T^0]}
If $a^* = a_0$ then at the end of $T$ rounds the following is true: 
$$\mathbb{E}[N^0_T] \geq T - \left( 2N(1+\frac{\pi^2}{3}) + \sum_{i,x} \frac{8\ln T}{\Delta_{i,x}^2} \right)~.$$
\end{lemma}
\begin{proof}
At the end of $T$ rounds we have
$$N^{0}_T + \sum_{i,x} N^{i,x}_T = T~. $$
Taking expectation on both sides of the above equation and rearranging the terms we have,
$$\mathbb{E}[N^{0}_T] = T -  \sum_{i,x} \mathbb{E}[N^{i,x}_T]~. $$
Now we use Lemma \ref{lemma: bound E[N_T^ix]} to conclude that
$$\mathbb{E}[N^{0}_T] \geq T - \left( 2N(1+\frac{\pi^2}{3}) + \sum_{i,x} \frac{8\ln T}{\Delta_{i,x}^2} \right)~. $$
\end{proof}

Now that we have bounds on $\mathbb{E}[N_T^0]$ and $\mathbb{E}[N_T^{i,x}]$, we can bound the regret as follows.

\textbf{Case a} ($a^* = a_0$): In this case we bound the expected cumulative regret of Algorithm \ref{CR-algorithm}. From Lemma \ref{lemma: bound E[N_T^ix]} and \ref{lemma: bound E[N_T^0]} for any $T$ satisfying both $T^{-\frac{p_{i,x}^2}{4}} > Z_i$ and 
\begin{align} \label{constraint on T}
    T \geq \frac{4}{p_{i,x} \cdot \eta_T^{i,x}} \bigg(1 + \frac{8\ln T}{\Delta^2_{i,x}} \bigg) + \bigg( 2N(1+\frac{\pi^2}{3}) + \sum_{i,x} \frac{8\ln T}{\Delta^2_{i,x}} \bigg)
\end{align}
we have $\mathbb{E}[N_T^{i,x}] \leq \frac{\pi^2}{3}$. Notice that Equation \ref{constraint on T} holds for sufficiently large $T$. Hence the cumulative regret caused by pulling sub-optimal arms $a_{i,x}$ is
\begin{align}
    \mathbb{E}[R(T)] \leq \sum_{\Delta_a > 0} \Delta_a \frac{\pi^2}{3}
\end{align}

\textbf{Case b} ($a^* \neq a_0$): In this case we bound the regret of pulling sub-optimal arms when $T \geq \max (L, e^{\frac{50}{\Delta_0^2}})$, where $L$ is as defined in Lemma \ref{bounding E[beta^2]}. Note that this is satisfied for sufficiently large $T$. Hence from Lemma \ref{bounding E[beta^2]} and Lemma \ref{lemma: bound E[N_T^ix]}, we have for $a^* \neq a_{i,x}$ and for $a_0$
\VIN{
\begin{align}
    \mathbb{E}[N_T^{i,x}] \leq \max \bigg(0, 1 + 8\ln T\bigg(\frac{1}{\Delta_{i,x}^2} - \frac{p_{i,x} \cdot \eta_T^{i,x}}{36\Delta_0^2} \bigg)\bigg) + \frac{\pi^2}{3}
\end{align}
\begin{align}
    \mathbb{E}[N_T^0] \leq \frac{58 \ln T}{\Delta_0^2} + 1 + \frac{\pi^2}{3}
\end{align}}

Hence the cumulative regret can be written as
\begin{align}
\mathbb{E}[R(T)] \leq \Delta_0 \bigg(\frac{58 \ln T}{\Delta_0^2} + 1 + \frac{\pi^2}{3}\bigg) + \sum_{\Delta_{i,x} > 0} \Delta_{i,x} \bigg(\max \bigg(0, 1 + 8\ln T\bigg(\frac{1}{\Delta_{i,x}^2} - \frac{p_{i,x} \cdot \eta_T^{i,x}}{36\Delta_0^2} \bigg)\bigg) + \frac{\pi^2}{3} \bigg)
\end{align}

\end{proof}

\section{Additional Experiments}
\label{secappendix:additional-experiments}
In this section, we provide two additional experiments. In Experiment $4$ we compare \SRM\ with a simple regret minimization algorithm \CB\ (referred to as Propagating Inference in figures), given in Algorithm $3$ in \cite{YabeHSIKFK18} and in Experiment $5$, we compare \CRM\ with \UCB\ in situations where observational arm is not the best arm.

 We mention few issues faced while implementing \CB\ using the details from \cite{YabeHSIKFK18} and how we resolved them: $(a)$ In Step $(3)$ of Algorithm $1$ in \cite{YabeHSIKFK18} (which is a subroutine for \CB), they iterate over all possible assignments to the parents of each node. Specifically, the algorithm would be exponential
time in the in-degree of the reward node $Y$ and therefore it runs efficiently only when $Y$ has a small number of parents. \SRM\ does not face this issue. To compare both algorithms we therefore created instances where in-degree of $Y$ was small. $(b)$ Another issue faced while implementing their algorithm is in an inequality condition specified in Equation $4$ of \cite{YabeHSIKFK18}. We observe that this inequality is trivially satisfied unless the time period becomes very large (of the order of $\geq 10^{10}$) even for their experiments given in Section $5$ of \cite{YabeHSIKFK18}. Since running the algorithms for such a long time period is not feasible, we run both algorithms till we see clear convergence of \SRM. $(c)$ A third problem we faced was in setting the time period range for our experiments. They use $T\in \{C, 2C, \ldots,9C\}$, but in Step $3$ of Algorithm $1$ and Step $4$ of Algorithm $2$ in \cite{YabeHSIKFK18}, they estimate probabilities using $T/3C$ samples. This would leave them with at most $3$ samples for such an estimation which would give noisy and unreliable estimates. Instead of using this set of values for $T$, we use equally spaced points in a time range where we see clear convergence of \SRM\ $(d)$ Finally, it is not discussed 
how the optimization problem giving $\hat{\eta}$ in Step $12$ of Algorithm $2$ of \cite{YabeHSIKFK18} is solved, and in the experiments they use a fixed value for $\hat{\eta}$. Since there is no technique proposed to solve the optimization problem, we also use the same fixed $\hat{\eta}$ that they used.

\begin{figure}[ht!]
    \centering
    \minipage{0.5\textwidth}
        \includegraphics[width=\columnwidth]{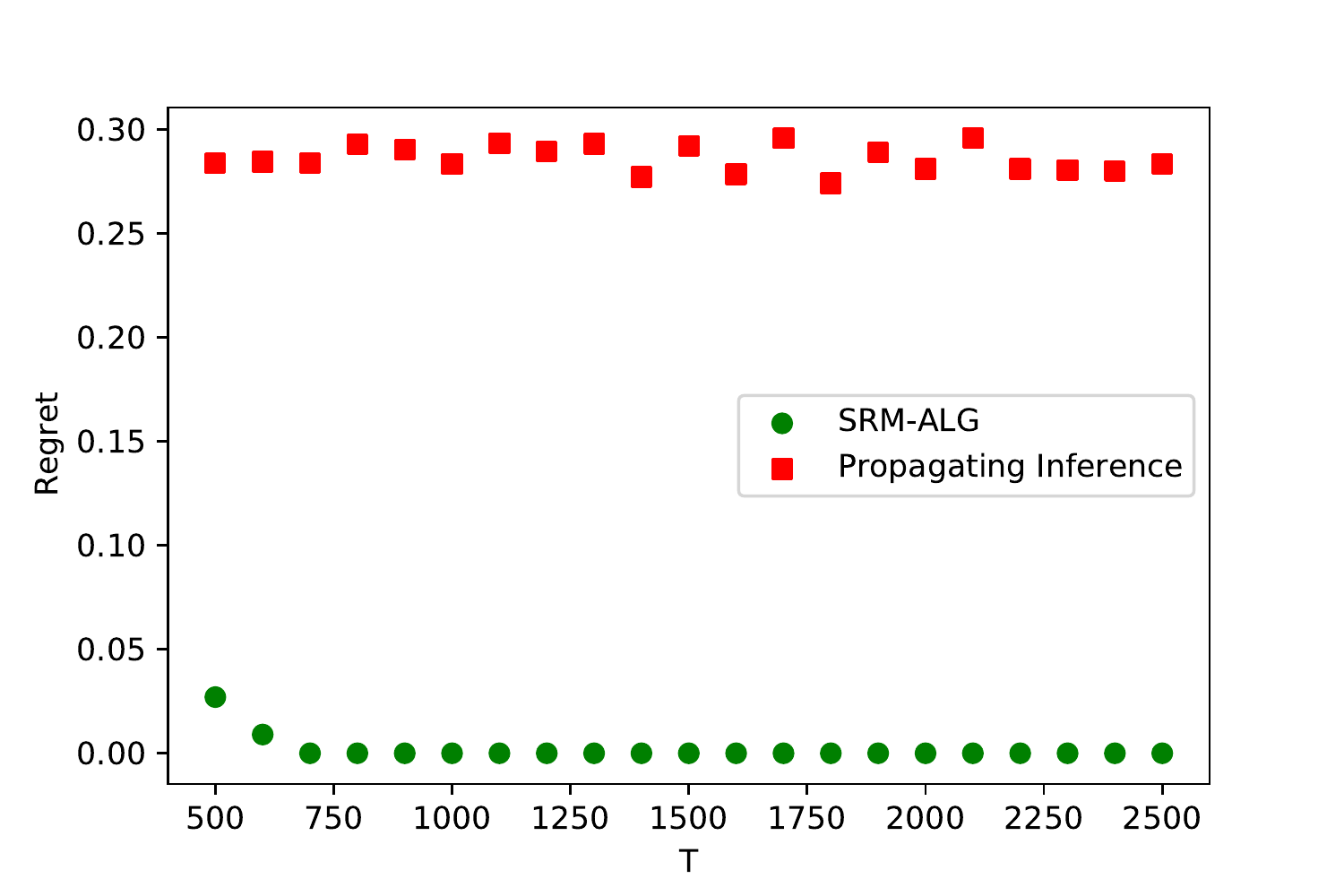}
        \caption{{\small Simple Regret vs T}}
        \label{fig:regret-vs-T_pi}
    \endminipage\hfill
    \minipage{0.5\textwidth}
      \includegraphics[width=\columnwidth]{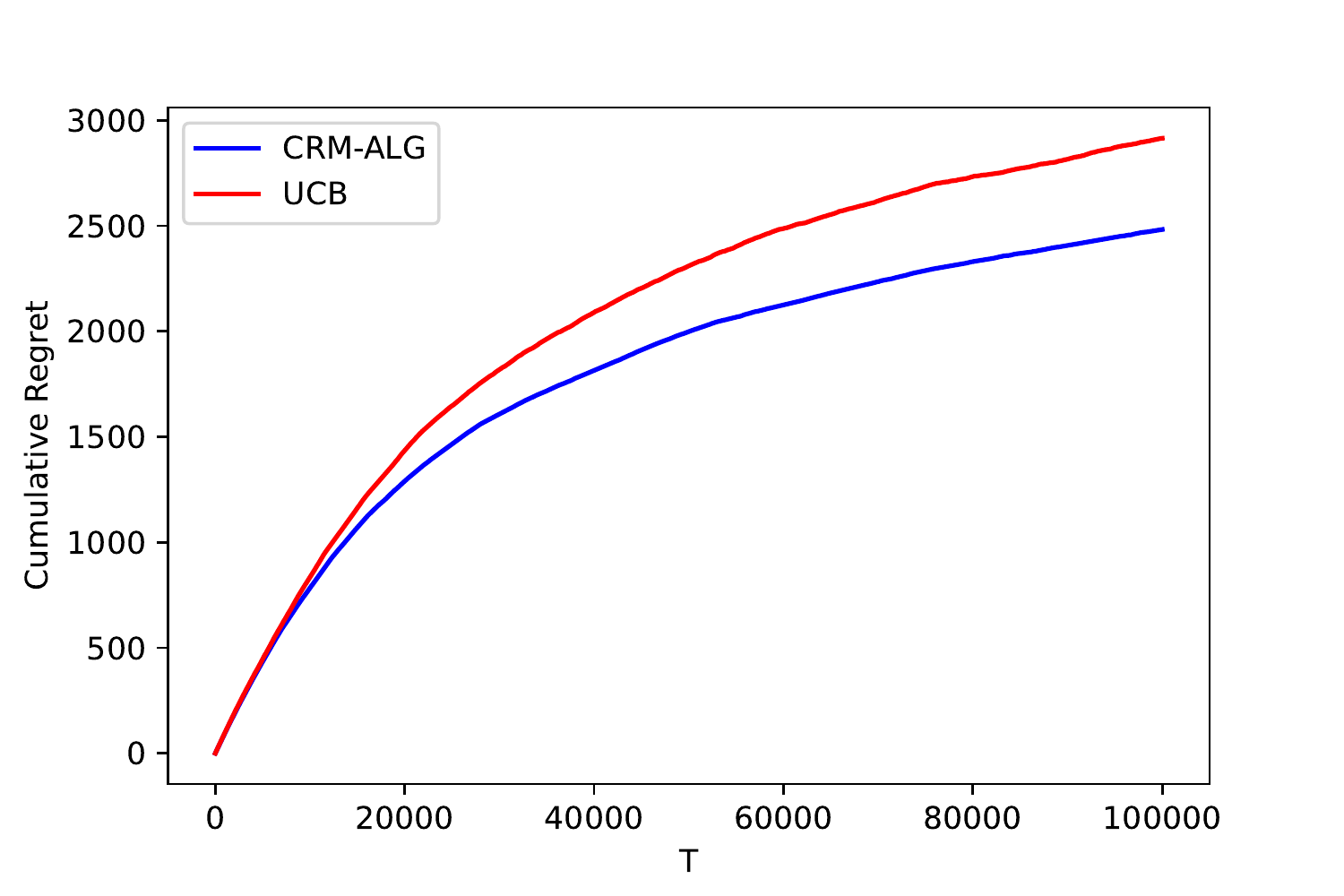}
        \caption[]{{\small Cumulative Regret vs T}} 	\label{fig:regret-vs-T-CRM-any-arm}   
    \endminipage\hfill
\end{figure}

\textbf{Experiment $4$ (Comparison with \cite{YabeHSIKFK18}):} This experiment compares the expected simple regret of \SRM\ with \CB\ as $T$ increases. We run the algorithms on 50 CBNs such that for every constructed CBN $\mathcal{C}$, it has $10$ intervenable nodes and $m(\mathcal{C})=5$. The CBNs are constructed as follows: a) randomly generate $50$ DAGs on $11$ nodes $X_1, \ldots,X_{10}$ and $Y$, and let $X_1\prec \ldots \prec X_{10}\prec Y$ be the topological order in each such DAG,\ b) $\mathbf{Pa}(X_i)$ contains at most $1$ node chosen uniformly at random from $X_1, \ldots, X_{i-1}$, and $\mathbf{Pa}(Y) = \{X_6, \dots, X_10\}$,\ c) $\mathbb{P}(X_i \mid \mathbf{Pa}(X_i)) = 0.5$ for $i\in [5]$ and $\mathbb{P}(X_i|\mathbf{Pa}(X_i)) = 1/10$ for $i\in [6,10]$,\ d) uniformly at random choose a $X_j$ from $\mathbf{Pa}(Y)$ and set the CPD of $Y$ as $\mathbb{P}(Y|\ldots, X_j=1,\ldots) = 0.5 + \epsilon$ and  $\mathbb{P}(Y|\ldots, X_j=0, \ldots) = 0.5 - \epsilon'$ where $\epsilon = 0.3$ and $\epsilon' = q\epsilon/(1-q)$ for $q=1/10$.
The choice of the conditional probability distributions (CPDs) in (c) ensures $m(\mathcal{C}) = 5$ for every CBN $\mathcal{C}$ that is generated. 
We note that the above strategy to generate CBNs is a generalization of of the one used in \cite{LattimoreLR16} to define parallel bandit instances with a fixed $m$.
For each of the $50$ random CBN, we run \SRM\ and \CB\ for multiple values of the time horizon $T$ in $[500, 2500]$ and average the regret over $30$ independent runs. Finally, we calculate the mean regret over all the $50$ random CBNs and plot mean regret vs. $T$ in Fig. \ref{fig:regret-vs-T_pi}. As seen in the Fig. \ref{fig:regret-vs-T_pi}, \SRM\ has a much lower regret as compared to \CB\ which incurs a regret of $\tilde{O}(\sqrt{N/T})$ in comparison to \SRM's regret of $\tilde{O}(\sqrt{m/T})$ that was proved in Theorem \ref{theorem: UB-SR}.

\textbf{Experiment $5$ (Cumulative Regret vs. T, General Case):} This experiment compares the cumulative regret of \CRM\ with \UCB\ as $T$ increases. The algorithms are run on 12 CBNs such that for every constructed CBN $\mathcal{C}$, it has $10$ intervenable nodes. The CBNs are constructed as follows: a) randomly generate $12$ DAGs on $11$ nodes $X_1, \ldots,X_{10}$ and $Y$, and let $X_1\prec \ldots \prec X_{10}\prec Y$ be the topological order in each such DAG,\ b) $\mathbf{Pa}(X_i)$ contains at most $1$ node chosen uniformly at random from $X_1, \ldots, X_{i-1}$, and $\mathbf{Pa}(Y)$ contains $X_i$ for all $i$,\ c) $\mathbb{P}(X_i \mid \mathbf{Pa}(X_i)) = 0.5$ for $i\in [10]$ and,\ d) uniformly at random choose a $X_j$ from $\mathbf{Pa}(Y)$ and set the CPD of $Y$ as $\mathbb{P}(Y|\ldots, X_j=1,\ldots) = 0.5 + \epsilon$ and  $\mathbb{P}(Y|\ldots, X_j=0, \ldots) = 0.5 - \epsilon'$ where $\epsilon = 0.1$ and $\epsilon' = q\epsilon/(1-q)$ for $q=1/2$, that is an interventional arm is the best arm. We run \CRM\ and \UCB\ for a sufficiently large $T$ and average the cumulative regrets over $30$ independent runs. Fig. \ref{fig:regret-vs-T-CRM-any-arm} demonstrates that cumulative regret of \CRM\ gets better compared to that of \UCB\ for large enough $T$ as expected by result in Theorem \ref{theorem: UB-CRM}.

\end{document}